\documentclass[letterpaper,twocolumn,10pt]{article}
\usepackage{usenix}
\usepackage{tikz}
\usepackage{amsmath}
\usepackage{filecontents}
\usepackage{booktabs}
\usepackage{multirow}
\usepackage{amsfonts}
\usepackage{enumitem}
\usepackage{xurl}
\usepackage{adjustbox}
\usepackage[most]{tcolorbox}
 
\usepackage{cleveref}
\usepackage{xcolor,pifont,xspace}
\usepackage[ruled,vlined]{algorithm2e}

\newcommand{\ournameNoSpace}{NeuronFuzz} 
\newcommand{\ourname}{\ournameNoSpace\xspace}

\newcommand{\oracleNoSpace}{SafetyOracle} 
\newcommand{\oracle}{\oracleNoSpace\xspace}

\newcommand{\dropval}[2]{%
  \ensuremath{#1\%_{\scriptscriptstyle\color{red}{-#2\%}}}%
}

\begin{document}

\title{\Large \bf \ourname: Safety Neuron Guided Fuzzing for LLM Safety Evaluation}

\author{
{\rm Zhiyuan Xu}\\
University of Bristol\\
zhiyuan.xu@bristol.ac.uk
\and
{\rm Muhammad Firhard Roslan}\\
University of Bristol\\
firhard.roslan@bristol.ac.uk
\and
{\rm Joseph Gardiner}\\
University of Bristol\\
joe.gardiner@bristol.ac.uk
\and
{\rm Sana Belguith}\\
University of Bristol\\
sana.belguith@bristol.ac.uk
\and
{\rm Lichao Wu}\\
University of Bristol\\
lichao.wu@bristol.ac.uk
}

% \author{
% {\rm Anonymous Author(s)}
% }

\maketitle

\begin{abstract}
Safety evaluation is critical for assessing whether aligned Large Language Models (LLMs) remain robust against jailbreak attacks. Existing automated testing methods, however, largely rely on response-level feedback: each candidate prompt typically requires generating a target-model response to evaluate its attack effectiveness. This process is expensive and, more importantly, provides only sparse guidance on strongly aligned models, where most candidates are rejected with the same failure outcome.

This paper presents NeuronFuzz, a white-box fuzzing framework that exploits internal safety neurons as continuous execution feedback for LLM safety evaluation. A SafetyOracle converts safety-neuron activations into a continuous safety alarm score that serves as feedback for fuzzing and can be obtained during prefill, eliminating response generation from the fuzzing loop. To construct the SafetyOracle, NeuronFuzz uses template-invariant harmful and benign inputs and stability-aware selection to identify a compact set of safety neurons whose activations capture harmful-intent recognition. Moreover, since the safety alarm score is differentiable, NeuronFuzz uses its gradients to identify safety-sensitive template positions and a masked language model to generate fluent, context-compatible mutations while preserving original harmful payload and avoiding additional optimization variables. We evaluate NeuronFuzz across 21 text and multimodal models. Across five white-box source models, it achieves a 76--100\% jailbreak discovery rate, outperforming baselines by up to 48 percentage points. Its optimized templates further transfer zero-shot to open-weight and six proprietary target models, achieving average ASR and top-5 ensemble ASR (EASR) of 69.6\%/92.6\% and 44.1\%/60.0\%, respectively.
\end{abstract}

\section{Introduction}
Large Language Models (LLMs) are increasingly deployed in applications that directly interact with users. To prevent misuse, modern LLMs undergo \emph{safety alignment} to recognize harmful requests and suppress unsafe responses. However, aligned models remain vulnerable to carefully crafted adversarial prompts, known as \emph{jailbreaks}, that bypass these safety mechanisms and elicit harmful content~\cite{wei2023jailbroken,shen2024anything}. As LLMs and their alignment strategies continue to evolve, systematically identifying such safety failures has become an important part of evaluating model robustness before deployment.

Existing LLM safety testing can broadly be divided into manual and automated approaches. Manual red teaming relies on human-designed strategies such as role-playing, instruction conflicts, refusal suppression, and contextual framing~\cite{hughes2026best,liu2023jailbreaking,wei2023jailbroken,shen2024anything}. While effective at exposing individual weaknesses, manual testing requires substantial human effort and domain expertise, limiting the scale and diversity of explored inputs. Automated approaches improve scalability by algorithmically generating or optimizing jailbreak prompts. For example, AutoDAN~\cite{liu2024autodan} evolves prompts through hierarchical genetic optimization; PAIR~\cite{chao2025jailbreaking} iteratively refines attacks using an attacker LLM. Another line of automated testing adapts fuzzing, a widely adopted technique for software and hardware testing~\cite{manes2019art,wu2025genhuzz}, to LLM safety evaluation~\cite{yu2024llm}. LLM fuzzers iteratively mutate jailbreak templates using language-based operators such as rephrasing, shortening, and expansion, and retain promising candidates based on response-level safety evaluation feedback. Through repeated mutation and selection, they can gradually discover more effective jailbreak templates. Despite their different search strategies, these methods follow the same evaluation paradigm: each candidate is submitted to the target LLM; a response is generated and then evaluated by a classifier, heuristic detector, or LLM judge.

However, this response-level paradigm introduces two fundamental limitations for automated safety testing: (i) \textbf{Sparse Search Feedback:}
Response-level evaluation provides little information about \emph{how promising} an unsuccessful candidate is. This problem is particularly severe for strongly aligned models, where most mutated prompts are refused and consequently receive the same failure outcome. Yet these candidates may affect the model's underlying safety behavior very differently: one may leave the safety mechanism largely unchanged, while another may substantially weaken it without yet producing a successful jailbreak. Response-level feedback cannot distinguish between them, leaving the fuzzer with limited guidance on which candidates to retain and further explore. Increasing the query budget does not fundamentally solve this problem because the feedback itself remains sparse.
(ii) \textbf{Expensive Response Generation:}
Evaluating every candidate requires autoregressive decoding, which is substantially more expensive than processing the input prompt alone~\cite{kamath2025pod, leviathan2023fast}. Response-based methods may additionally invoke external classifiers or LLM judges after generation. Since automated fuzzers repeatedly evaluate large numbers of candidates, complete response generation becomes a major scalability bottleneck. Thus, an efficient fuzzer requires a feedback signal that can evaluate candidates \emph{before response generation}.

\noindent
\textbf{Our Approach and Contributions:}
To address these limitations, we present \ourname, a white-box fuzzing framework that uses internal safety-neuron activations as continuous execution feedback. Inspired by coverage-guided fuzzing~\cite{zalewski2014}, \ourname constructs a lightweight \oracle that maps a compact set of stable safety neurons to a continuous safety alarm score during prefill. To ensure that this signal captures harmful intent rather than jailbreak-template artifacts, we identify safety neurons using template-invariant harmful--benign pairs and stability-aware selection.
The \oracle eliminates response generation for intermediate candidates and distinguishes promising mutations even when they share the same response-level failure outcome. Moreover, its differentiable score allows \ourname to locate safety-sensitive template positions through gradients, while a masked language model generates context-compatible replacements that preserve the harmful payload and natural-language structure. Our main contributions are as follows:

% \begin{itemize}[itemsep=0pt, parsep=0pt, topsep=0pt, leftmargin=*]
\begin{itemize} [itemsep=0pt, topsep=0pt, leftmargin=*]
    \item We introduce a new perspective on LLM fuzzing that uses \emph{safety neurons as execution feedback}, replacing expensive and sparse response-level evaluation with continuous internal feedback available during prefill.
    \item We design a lightweight \oracle based on template-invariant activation extraction and stability-aware safety-neuron selection. Its differentiable safety alarm score guides both candidate ranking and gradient-based localization of safety-sensitive template positions.
    \item We develop \ourname, which combines prefill-only \oracle feedback with gradient-guided masked-token mutation to efficiently explore jailbreak templates while preserving the harmful payload and natural-language structure. 
    \item We conduct extensive evaluations across 21 text and multimodal models. \ourname{} consistently improves jailbreak discovery over representative baselines and its optimized templates transfer across model families, proprietary APIs, datasets, and visual safety tasks.
\end{itemize}

The remainder of this paper is organized as follows. Section~\ref{sec:Preliminaries} provides background on LLM safety evaluation and fuzzing. Section~\ref{sec:Method} presents the motivation, threat model, and design of \oracle and \ourname. Section~\ref{sec:Experiment Setup} describes the implementation and experimental setup. Section~\ref{sec:Experimental Results} reports the main results, and Section~\ref{sec:Ablation and Hyperparameter Study} presents the ablation studies. 
Section~\ref{sec:Defenses} evaluates input defenses and defensive uses of the \oracle{}. 
% Finally, Sections~\ref{sec:Discussion} and~\ref{sec:Conclusion} discuss the findings and conclude the paper.
Finally, Section~\ref{sec:Conclusion} concludes the paper.

\section{Preliminaries}
\label{sec:Preliminaries}

\subsection{LLM Safety Evaluation}
\label{subsec:Attacks on LLMs}

Aligned large language models (LLMs) are trained to refuse harmful requests, but their safety robustness still needs to be systematically evaluated before and after deployment. LLM safety evaluation examines whether a model continues to follow its safety policy when exposed to challenging or adversarial inputs~\cite{wei2023jailbroken,shen2024anything}. A common jailbreak test consists of a harmful payload and a jailbreak template~\cite{googledeepmind2026gemini3pro,openai2026gpt56previewchallenging}. The payload specifies the harmful request, while the template changes how the request is presented to the model. The same template can be combined with different payloads to examine whether a safety weakness generalizes across harmful tasks. Existing methods can be broadly grouped into three approaches: manual testing, automated testing, and internal safety analysis. 

\noindent\textbf{Manual Testing.} 
Early LLM safety testing mainly relied on manually designed adversarial prompts. Human testers construct prompts using strategies such as role-playing, fictional scenarios, instruction conflicts, refusal suppression, and emotional framing to probe weaknesses in model alignment~\cite{shen2024anything,liu2023jailbreaking,wei2023jailbroken,hughes2026best}. These tests are easy to interpret and show that even small changes in prompt context may substantially affect model behavior~\cite{andriushchenko2025jailbreaking}.  However, manual testing requires substantial human effort and domain knowledge, which limits the number and diversity of test cases that can be explored.

\noindent\textbf{Automated Safety Testing.}
Automated methods reduce this manual effort by generating or refining jailbreak prompts algorithmically. Different methods adopt different optimization strategies. AutoDAN uses a hierarchical genetic algorithm to evolve jailbreak prompts at sentence and word levels~\cite{liu2024autodan}; PAIR uses an attacker LLM to iteratively refine prompts based on target-model responses and an external judge~\cite{chao2025jailbreaking}; and LLM-Fuzzer automatically generates variants of existing jailbreak templates~\cite{yu2024llm}. Although their optimization procedures differ, these methods share the goal of producing effective safety test cases with less human intervention. Their effectiveness is typically determined from the target model's generated response, providing an external measure of whether a test case successfully bypasses the model's safety policy.

\noindent\textbf{Internal Safety Analysis.}
Beyond model outputs, prior studies show that safety-related information is also represented internally. Harmful and benign inputs can produce different activation patterns, while refusal behavior can be associated with specific representation directions or internal components~\cite{turner2023steering,arditi2024refusal,park2023linear,rimsky2024steering,xu2025steering}. More recent neuron-level studies further show that this information can be localized to compact neuron subsets whose activations selectively distinguish harmful from benign inputs~\cite{wu2025neurostrike,kazemi2026single}. Manipulating these neurons or related internal components can substantially alter model's safety behavior, providing evidence that they are functionally related to safety and refusal~\cite{wu2025neurostrike,kazemi2026single,xu2026routehijack,wu2025gatebreaker,thang2026goodvibe}. These findings suggest that internal activations expose safety-related information that is not directly observable from the final response. Our work builds on this observation and uses safety-related neuron activations as signals for automated safety testing without modifying the model's internal states.

\subsection{Fuzzing}
\label{subsec:Fuzzing}

Fuzzing is an automated testing technique that repeatedly generates random test inputs to expose unexpected program behaviors and security vulnerabilities~\cite{manes2019art}. Fuzzing methods can be broadly classified as black-box, grey-box, or white-box according to the amount of information available from the target program. A coverage-guided greybox fuzzing process~\cite{zalewski2014,Dechand2026} typically contains four components: a seed corpus and scheduler~\cite{Chen2018}, an input mutator~\cite{Lemieux2018}, runtime instrumentation~\cite{zalewski2014}, and execution feedback~\cite{Bohme2016}. The fuzzing process begins with a corpus of initial seed inputs. In each iteration, the scheduler assigns energy values to seeds according to a scheduling strategy. Seed inputs that exercise new coverage goals based on the feedback receive higher energy and are more likely to be selected for next mutation. The mutator then modifies the selected seed to generate new tests, which are executed by the target program. An oracle~\cite{barr2014oracle} then determines whether the execution reveals an unexpected behavior (e.g., the program crashes). By repeatedly using execution information to update seed selection and mutation, the fuzzer can progressively explore the program's execution space.

LLM fuzzing adapts this feedback-guided process to natural-language safety testing~\cite{yu2024llm}. Jailbreak templates can serve as initial seeds, while language-based mutation operators generate new template variants through operations such as expansion, rephrasing, or shortening. The target program is an LLM, and a successful test corresponds to a generated response that violates the intended safety policy. 
% Unlike traditional software, however, LLMs do not expose explicit crashes or execution coverage through a standard input-output interface. 
% Existing LLM fuzzers therefore evaluate generated responses 
The generated responses are evaluated
using classifiers, heuristic rules, or LLM judges, which serve as response-level oracles~\cite{chao2025jailbreaking,yu2024llm, liu2024autodan}. The evaluation outcomes are then used to decide which prompts should be retained and explored further. 

However, this response-level paradigm has two limitations for LLM safety evaluation. First, each prompt must generate a response before it can be evaluated, introducing repeated decoding and evaluation overhead. Second, the resulting feedback provides limited information for distinguishing prompts with the same response-level outcome. This problem becomes more pronounced on strongly aligned models, where many mutated prompts are rejected and receive the same failure label. The fuzzer therefore has little information about which unsuccessful prompts are more promising for further exploration. These limitations motivate a feedback signal that can evaluate and distinguish prompts before response generation.

\section{\ourname}
\label{sec:Method}

\subsection{Motivation}
\label{subsec:Motivation}

\begin{figure*}[ht]
  \centering
  \includegraphics[width=0.95\linewidth]{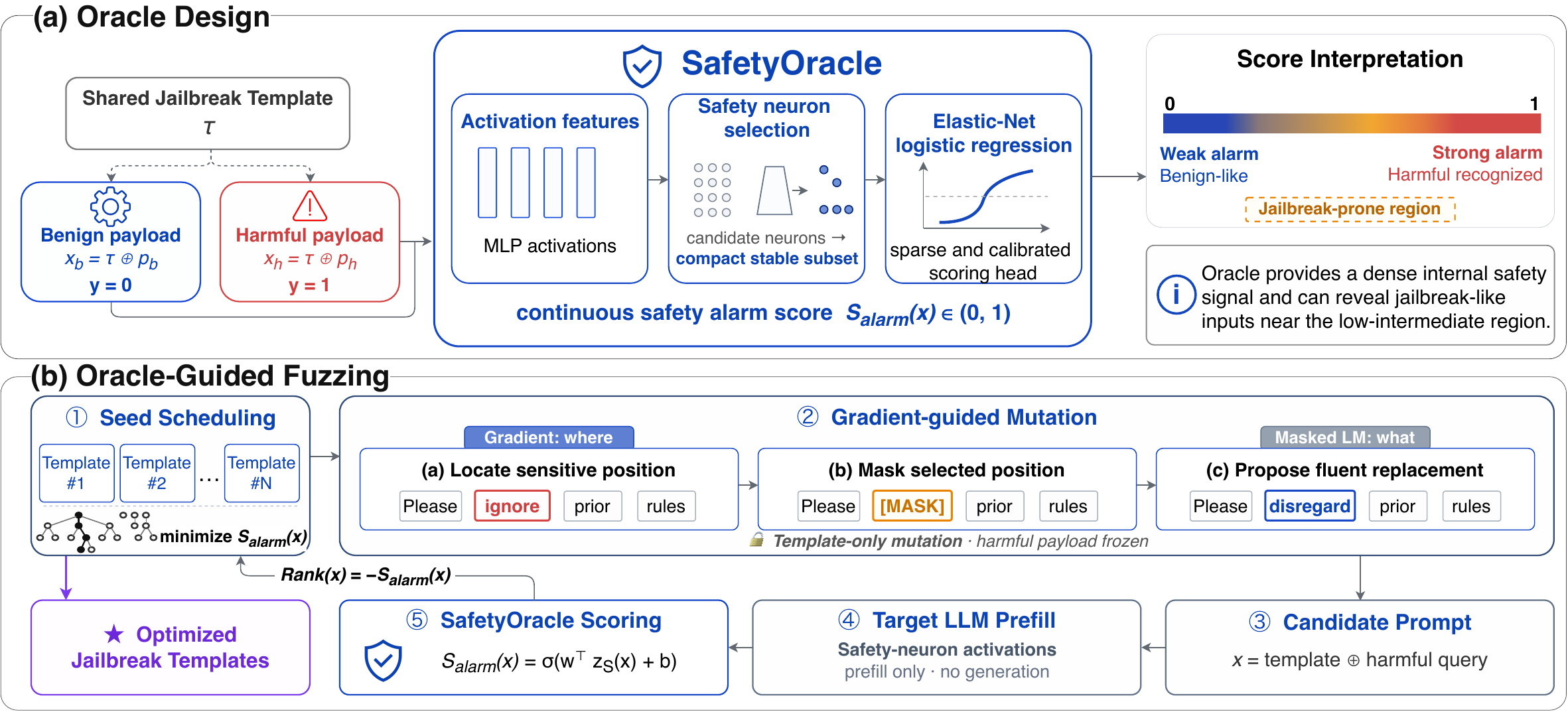}
  \caption{Overview of \ourname{}. (a) The \oracle{} extracts MLP activations from template-invariant harmful and benign pairs, selects stable safety neurons, and produces a continuous safety alarm score. (b) The fuzzer minimizes this score through gradient-guided, template-only mutation and evaluates candidates using prefill-only oracle feedback.}
  \label{img:main}
\end{figure*}

\noindent\textbf{From Response-Level Feedback to Internal Safety Signals.}
Our key intuition is to use internal safety representations as a denser source of fuzzing feedback. Prior white-box studies show that safety-related information is encoded in model activations and can be captured by a small set of internal features~\cite{wu2025neurostrike,kazemi2026single}. Rather than modifying these representations, we use their activations only as signals to guide safety testing. Internal safety signals have three properties that make them suitable for fuzzing: (i) they can be obtained during the prefill stage without generating a response, reducing the cost of evaluating intermediate candidates; (ii) they provide continuous information that can distinguish candidates with the same response-level outcome. This is particularly useful on strongly aligned models, where many mutated prompts are rejected but may still produce different internal safety signals; (iii) these signals remain differentiable with respect to the input, allowing gradients to identify token positions that strongly affect the model's safety representation. 

\noindent\textbf{Stable and Intent-Focused Feedback.}
Raw activations, however, cannot be directly used as reliable fuzzing feedback. The activation space is high-dimensional and contains noisy or redundant features. Jailbreak templates may also introduce activation patterns that reflect prompt style rather than the harmful intent of the payload. We therefore seek a compact and stable signal that captures harmful intent while remaining robust to jailbreak contexts. To achieve this, we construct template-invariant harmful and benign pairs and apply stability-aware neuron selection for the training of \oracle{}.

\noindent\textbf{Gradient-Guided Mutation.}
A useful fuzzing signal should not only rank candidates but also guide how they are mutated. Beyond providing continuous feedback for candidate ranking, the \oracle{} is differentiable with respect to the input, allowing us to identify which template positions most strongly influence the internal safety signal. We therefore design a gradient-guided mutation strategy that uses gradients to locate sensitive positions and a masked language model to generate context-compatible replacements. In this design, the gradient determines \emph{where} to mutate, while the masked language model determines \emph{what} to replace. We restrict all mutations to the jailbreak template and keep the harmful payload unchanged. Together, these designs allow \ourname{} to use internal safety signals for both continuous candidate ranking and targeted prompt mutation.
Figure~\ref{img:main} summarizes the two stages of \ourname{}. Figure~\ref{img:main}(a) presents the construction of the \oracle{} and the interpretation of its safety alarm score. Figure~\ref{img:main}(b) shows how \oracle is integrated into the online fuzzing loop: gradients localize sensitive template positions, the masked language model generates replacements, and the resulting candidates are scored during prefill to guide subsequent seed selection.

\subsection{Threat Model}
\label{sec:ThreatModel}

We consider two settings for \ourname{}: an authorized evaluator and a malicious adversary. The evaluator uses white-box information to identify safety weaknesses, while the adversary uses a surrogate model to optimize jailbreak templates and applies them either to the same model or transfers them to other target models.

\noindent\textbf{Evaluator.}
The evaluator may be the model developer, service provider, or an authorized third-party auditor. The evaluator has access to the model's parameters, MLP activations, and input gradients, which are used to construct the \oracle{} and guide template mutation. The evaluator does not modify model parameters, internal activations, system prompts, or decoding settings. The discovered test cases are therefore evaluated through the model's normal input interface.

\noindent\textbf{Adversary.}
We follow the prompt-level jailbreak setting used by prior automated jailbreak methods~\cite{zou2023universal,liu2024autodan,yu2024llm}. The adversary has white-box access to a local source or surrogate model and uses \ourname{} to optimize jailbreak templates. These templates can attack the source model directly or be transferred zero-shot to other target models. In the transfer setting, the adversary requires only input-output access to the target model.

\subsection{\oracle Design}
\subsubsection{Template-invariant Activation Extraction}
\label{subsec:Template-invariant Activation Extraction}

Safety-related neurons are identified by contrasting the activations of harmful and benign prompts~\cite{wu2025neurostrike,turner2023steering}. The goal is to extract features that primarily reflect harmful intent and remain stable across different prompt contexts. However, jailbreak templates may introduce additional activation patterns through role-playing instructions, refusal suppression, or unusual linguistic structures. These template-specific patterns can act as shortcuts, causing the extracted features to reflect template style rather than payload harmfulness. To reduce this effect, we construct template-invariant harmful--benign pairs that share the same jailbreak template but contain payloads with different safety labels. Let $\mathcal{T}=\{\tau_1,\ldots,\tau_M\}$ denote the jailbreak templates, and let $\mathcal{P}_{h}$ and $\mathcal{P}_{b}$ denote the harmful and benign payload sets. Given a template $\tau$ and payload $p$, the model input is $x=\tau\oplus p$, where $\oplus$ inserts the payload into the template. Each template therefore forms a paired input:
\begin{equation}
x^{h}_{i,j}=\tau_j\oplus p^{h}_{i},
\qquad
x^{b}_{i,j}=\tau_j\oplus p^{b}_{i},
\label{eq:template_invariant_pair}
\end{equation}
where $p^{h}_{i}\in\mathcal{P}_{h}$ and $p^{b}_{i}\in\mathcal{P}_{b}$, with labels $y(x^{h}_{i,j})=1$ and $y(x^{b}_{i,j})=0$. Since the same template appears in both classes, the label difference is determined by the payload rather than the surrounding template. For each input, a prefill pass is performed while forward hooks collect the outputs of the gate and up projections in every Transformer MLP block~\cite{wu2025gatebreaker,thang2026goodvibe}. Let $\mathcal{I}(x)$ denote the input-prompt token positions. For layer $\ell$ and projection module $m\in\{\mathrm{gate},\mathrm{up}\}$, token-level outputs $\mathbf{z}^{(\ell,m)}_t(x)$ are aggregated by element-wise max pooling:

\begin{equation}
\mathbf{a}^{(\ell,m)}(x)
=
\max_{t\in\mathcal{I}(x)}
\mathbf{z}^{(\ell,m)}_t(x).
\label{eq:activation_pooling}
\end{equation}

This produces a fixed-dimensional activation vector for each layer and projection module, independent of prompt length. The pooled vectors are concatenated across all layers and modules to obtain the prompt-level representation $\mathbf{a}(x)$, with each dimension treated as a candidate neuron feature. Section~\ref{ablation:Neuron Selection} ablates the design of template-invariant extraction.

\subsubsection{Stability-Aware Safety Neuron Selection}
\label{subsec:Stability-Aware Safety Neuron Selection}
Template-invariant extraction reduces template-specific effects, but neuron importance estimated from a single training set can still be sensitive to sampling variation~\cite{kim2018interpretability, meinshausen2010stability}. A neuron that appears strongly predictive in one sample may become weak or change its association in another~\cite{kim2018interpretability}. We therefore use bootstrap-based stability selection to identify neurons that remain consistently associated with harmful inputs across resampled training sets~\cite{meinshausen2010stability}. We perform $B$ bootstrap runs and fit a linear classifier in each run. For an MLP block $\ell$ containing $p_\ell$ candidate neurons, we retain the $q_\ell$ neurons with the largest absolute coefficients, where
$q_\ell=\left\lceil \sqrt{p_\ell} \right\rceil$.
This provides a consistent selection size across models with different hidden dimensions.

For each neuron $j$, let $n_j^{+}$ and $n_j^{-}$ denote the numbers of runs in which it is retained with a positive or negative coefficient. We define its selection rate $s_j$ and sign consistency $c_j$ as

\begin{equation}
\begin{aligned}
s_j &=
\frac{n_j^{+}+n_j^{-}}{B},
&
c_j &=
\frac{\max(n_j^{+},n_j^{-})}
{n_j^{+}+n_j^{-}}.
\end{aligned}
\label{eq:stability_selection}
\end{equation}

We set $c_j=0$ if neuron $j$ is never retained. Here, $s_j$ measures how consistently the neuron is selected, while $c_j$ measures whether its coefficient maintains the same direction across bootstrap runs. Since harmful inputs are assigned the positive label, we further require the dominant coefficient direction to be positive. Let $\tau_s$ and $\tau_c$ denote the thresholds for selection rate and sign consistency, respectively. The final safety-neuron set is

\begin{equation}
\mathcal{S}
=
\left\{
j \,\middle|\,
s_j \geq \tau_s,\;
c_j \geq \tau_c,\;
n_j^{+} > n_j^{-}
\right\}.
\label{eq:safety_neuron_set}
\end{equation}

This selection removes neurons that are unstable across resampled training sets or whose association with harmful inputs changes direction. The resulting activation vector $\mathbf{a}_{\mathcal{S}}(x)$ is used as the input to the \oracle. Section~\ref{ablation:Neuron Selection} compares our method with prior statistic-based neuron selection.

\subsubsection{Neuron-Based \oracle}
\label{subsec:Neuron-Based SafetyOracle}

The selected safety neurons capture safety-related information, but their activations form a multidimensional representation rather than a single score that can be directly compared across candidate prompts. We therefore train a lightweight classifier as \oracle to map their joint activation pattern into a continuous safety signal~\cite{du2024haloscope,burger2024truth}. Given an input $x$, let $\mathbf{a}_{\mathcal{S}}(x)$ denote the activation vector of the selected safety neurons. We standardize each dimension as $\mathbf{z}_{\mathcal{S}}(x)=(\mathbf{a}_{\mathcal{S}}(x)-\boldsymbol{\mu})/\mathbf{s}$, where $\boldsymbol{\mu}$ and $\mathbf{s}$ are the training-set mean and standard deviation. The \oracle{} computes:

\begin{equation}
S_{\mathrm{alarm}}(x)
=
\operatorname{sigmoid}
\left(
\mathbf{w}^{\top}\mathbf{z}_{\mathcal{S}}(x)+b
\right)
\in (0,1).
\label{eq:safety_oracle}
\end{equation}

We refer to $S_{\mathrm{alarm}}(x)$ as the safety alarm score. Since harmful inputs are assigned the label $y=1$, a higher score indicates an activation pattern that is more characteristic of harmful prompts, suggesting stronger internal recognition of harmful intent. A lower score indicates a weaker internal safety alarm.

Although proposed stability selection in Section~\ref{subsec:Stability-Aware Safety Neuron Selection} removes many unreliable neurons, the retained features may still contain redundant or correlated information~\cite{meinshausen2010stability}. We therefore implement the \oracle{} using logistic regression with Elastic Net regularization~\cite{friedman2010regularization}. Elastic Net combines $\ell_1$ and $\ell_2$ regularization. The $\ell_1$ term encourages uninformative feature weights to become zero, while the $\ell_2$ term stabilizes the weights of correlated features. This combination allows the \oracle{} to further control redundancy in the selected neuron features when learning the final scoring function.

Let
$\mathcal{L}_{\mathrm{cls}}
=
\frac{1}{N}\sum_{i=1}^{N}
\mathcal{L}_{\mathrm{BCE}}(y_i,S_{\mathrm{alarm}}(x_i))$
denote the classification loss based on the $S_{\mathrm{alarm}}$ in Eq.~\ref{eq:safety_oracle}. We optimize

\begin{equation}
\min_{\mathbf{w},b}
\mathcal{L}_{\mathrm{cls}}
+
\frac{1}{NC}
\left[
\rho\lVert\mathbf{w}\rVert_1
+
\frac{1-\rho}{2}\lVert\mathbf{w}\rVert_2^2
\right],
\label{eq:elastic_net_oracle}
\end{equation}
where $N$ is the number of training examples, $C>0$ controls the inverse regularization strength, and $\rho\in[0,1]$ balances the $\ell_1$ and $\ell_2$ penalties. We compare Elastic Net with four other classification models and justify this choice in Appendix~\ref{appendix:oracle-model-selection}. During fuzzing, the \oracle{} is used only to guide candidate selection and mutation. Each candidate requires only a prefill pass and does not require the target model to generate the response. The continuous safety alarm score $S_{\mathrm{alarm}}$ allows the fuzzer to rank candidates that would otherwise receive the same response-level failure label. Moreover, because the \oracle{} remains differentiable with respect to the input embeddings, its gradients can identify sensitive token positions for the mutation process described in Section~\ref{subsection:Gradient-Guided Mutation}.

\subsection{\ourname{} Pipeline}
\label{subsec:NeuronFuzz Pipeline}

Following the software fuzzing described in Section~\ref{subsec:Fuzzing}, \ourname{} adapts the same feedback-guided pipeline to safety evaluation. In each iteration, the scheduler first selects a jailbreak template for mutation. The mutator then uses \oracle{} gradients to identify sensitive template positions and a masked language model to generate context-compatible replacements. Each resulting candidate is executed through a prefill pass to collect the selected safety-neuron activations. Finally, the \oracle{} converts these activations into a safety alarm score and uses it to update the seed scheduler for the next iteration. The workflow is shown in Figure~\ref{img:main}(b).

\subsubsection{Seed Scheduler}
The seed scheduler determines which templates are selected for further mutation. We follow prior work and adopt MCTS-Explore~\cite{yu2024llm}, where each node represents a jailbreak template and each edge records the mutation relationship between a parent and its child. The scheduler selects templates according to the statistics accumulated from previous oracle feedback, balancing the exploration of new templates with the reuse of promising ones. The selected template is then passed to the gradient-guided mutator.

\subsubsection{Gradient-Guided Mutation}
\label{subsection:Gradient-Guided Mutation}

Given a selected template $\tau$ and a harmful payload $p_h$, we construct the input $x=\tau\oplus p_h$. Rather than mutating template tokens uniformly or directly optimizing arbitrary token replacements~\cite{zou2023universal}, we use the \oracle{} to focus mutation on positions that most strongly influence the internal safety signal. Specifically, \oracle{} gradients determine \emph{where} to mutate, while a masked language model determines \emph{what} context-compatible replacement to use. This design narrows the mutation space through token replacement, preserving the prompt length and natural-language structure without introducing additional tokens or auxiliary optimization variables. 

Since the \oracle{} is differentiable with respect to the input, we backpropagate the safety alarm score to the input embeddings. Let $\mathbf{e}_t$ denote the embedding of token $t$. Its sensitivity is defined as:

\begin{equation}
g_t
=
\left\|
\frac{\partial S_{\mathrm{alarm}}(x)}
{\partial \mathbf{e}_t}
\right\|_2.
\label{eq:token_sensitivity}
\end{equation}

A larger $g_t$ indicates that the corresponding position has a stronger influence on the internal safety signal. We therefore rank template tokens by $g_t$ and select the top-ranked positions for mutation. For each selected position, the original token is masked and a lightweight masked language model proposes context-compatible replacements. Each replacement produces a new template $\tau'$, which is combined with the unchanged payload to form $x'=\tau'\oplus p_h$. All mutations are restricted to the jailbreak template, ensuring that a successful test reflects a safety bypass rather than a modification of the underlying payload (i.e., harmful request)~\cite{souly2024strongreject,salinas2024butterfly}.

\subsubsection{Runtime Instrumentation}
For each candidate input $x'$, \ourname{} performs a prefill pass through the target model and collects the activations of the selected safety-neuron set $\mathcal{S}$. These activations serve as the runtime execution signal and are passed to the \oracle{} for evaluation. Unlike existing response-level fuzzing~\cite{yu2024llm,chao2025jailbreaking}, this step does not require autoregressive response generation.

\subsubsection{Oracle Feedback and Seed Update}
Given the collected safety-neuron activations, the \oracle{} computes the safety alarm score $S_{\mathrm{alarm}}(x')$ using Eq.~\ref{eq:safety_oracle}. Since a lower score indicates weaker internal recognition of harmful intent, \ourname{} defines the fuzzing reward as $R(x')=-S_{\mathrm{alarm}}(x').$ The candidate and its reward are then added to the MCTS tree to update the scheduler. Candidates with lower safety alarm scores receive higher rewards and are therefore more likely to be selected for subsequent mutation. Compared with prior response-level feedback, $S_{\mathrm{alarm}}$ provides a continuous signal, allowing the scheduler to distinguish candidates even when they would receive the same response-level outcome. This feedback closes the fuzzing loop and guides subsequent exploration toward more promising templates.
\section{Implementation \& Experiment Setup}
\label{sec:Experiment Setup}
\begin{table*}[!ht]
\centering
\caption{Specifications of target LLMs.}
\label{tab:models}
\small
\setlength{\tabcolsep}{4pt}
\resizebox{\textwidth}{!}{
\begin{tabular}{l|c|cccc|cc}
\toprule
\textbf{Model} &
\textbf{Usage} &
\textbf{Architecture} &
\textbf{Active/Total Params (B)} &
\textbf{Reasoning} &
\textbf{Access} &
\textbf{Release Date} &
\textbf{Provider} \\
\midrule

DeepSeek-R1-14B~\cite{guo2025deepseek}
& Source
& Dense
& 14.7 / 14.7
& CoT
& White-box
& 2025.01
& DeepSeek \\

GPT-OSS-20B~\cite{agarwal2025gpt}
& Source
& MoE
& 3.6 / 21.0
& CoT
& White-box
& 2025.08
& OpenAI \\

Gemma-3-4B-it~\cite{gemmateam2025gemma3}
& Source
& Dense
& 4.0 / 4.0
& Non-CoT
& White-box
& 2025.03
& Google \\

Gemma-4-E4B-it~\cite{gemma4technicalreport}
& Source
& Dense
& 4.5 / 8.0$^{\dagger}$
& CoT/Non-CoT
& White-box
& 2026.04
& Google \\

Llama-3.1-8B-Instruct~\cite{grattafiori2024llama}
& Source
& Dense
& 8.0 / 8.0
& Non-CoT
& White-box
& 2024.07
& Meta \\

Llama-3.2-3B-Instruct~\cite{meta2024llama32}
& Transfer
& Dense
& 3.0 / 3.0
& Non-CoT
& Inference only
& 2024.09
& Meta \\

Phi-4~\cite{abdin2024phi}
& Transfer
& Dense
& 14.0 / 14.0
& Non-CoT
& Inference only
& 2024.12
& Microsoft \\

Qwen3.6-35B-A3B~\cite{qwen36_35b_a3b}
& Transfer
& MoE
& 3.0 / 35.0
& CoT/Non-CoT
& Inference only
& 2026.04
& Alibaba \\

Qwen3.6-27B~\cite{qwen3.6-27b}
& Transfer
& Dense
& 27.0 / 27.0
& CoT/Non-CoT
& Inference only
& 2026.04
& Alibaba \\

GLM-4.7-Flash~\cite{5team2025glm45agenticreasoningcoding}
& Transfer
& MoE
& 3.0 / 30.0
& CoT/Non-CoT
& Inference only
& 2026.01
& Z.ai \\

\midrule

LongCat-2.0~\cite{meituan2026longcat20}
& Transfer
& MoE
& 48.0 / 1600.0
& CoT/Non-CoT
& API inference only
& 2026.06
& Meituan \\

DeepSeek-V4-Flash~\cite{xu2026deepseek}
& Transfer
& MoE
& 13.0 / 284.0
& CoT/Non-CoT
& API inference only
& 2026.04
& DeepSeek \\

DeepSeek-V4-Pro~\cite{xu2026deepseek}
& Transfer
& MoE
& 49.0 / 1600.0
& CoT/Non-CoT
& API inference only
& 2026.04
& DeepSeek \\

Grok-4.5~\cite{xai2026grok45}
& Transfer
& MoE
& Undisclosed
& CoT
& API inference only
& 2026.07
& SpaceXAI \\

Gemini-2.5-Pro~\cite{comanici2025gemini}
& Transfer
& Undisclosed
& Undisclosed
& CoT
& API inference only
& 2025.06
& Google \\

Gemini-3-Flash~\cite{googledeepmind2025gemini3flash}
& Transfer
& Undisclosed
& Undisclosed
& CoT/Non-CoT
& API inference only
& 2025.12
& Google \\

\bottomrule
\end{tabular}
}

\footnotesize
\raggedright
$^{\dagger}$ For Gemma-4-E4B-it, 4.5B denotes the effective parameter count, whereas 8.0B includes the per-layer embedding parameters. CoT/Non-CoT denotes models that support both explicit thinking and direct-answer modes. We disable CoT for hybrid models by default for evaluation efficiency. 
\end{table*}

\subsection{Fuzzing Details}
\label{subsec:Implementation Details}

\noindent\textbf{Activation Extraction and Neuron Selection.} We register forward hooks on the gate and up projection modules of every Transformer block and apply element-wise max pooling~\cite{wu2025neurostrike, wu2025gatebreaker}. To remove the noise introduced by chat format, we retain only the activation positions that correspond to the input prompt. For stability selection, we perform $B=100$ bootstrap runs. Following previous research~\cite{khaire2022stability}, each run samples 70\% of the training examples and fits a logistic regression classifier on the extracted activations. A neuron is included in the final set when its selection rate is at least 0.6, its sign consistency is at least 0.75, and its dominant coefficient direction is positive~\cite{meinshausen2010stability}.

\noindent\textbf{SafetyOracle Training.} We standardize each selected neuron using the mean and standard deviation computed from the training set. We train one \oracle for each source model using Elastic Net logistic regression. We set the inverse regularization strength to $C=1.0$ and the $\ell_1$ ratio to $\rho=0.5$. We use the SAGA solver with a maximum of 5,000 iterations and a convergence tolerance of $10^{-4}$. 

\noindent\textbf{Mutation.}
For gradient-guided mutation, we first identify three template token positions with the largest gradient norms using Eq.~\ref{eq:token_sensitivity}. We use ModernBERT-base (149M parameters) as the masked language model for token replacement~\cite{modernbert}. At each selected position, we replace the original token with a mask token and use ModernBERT to propose context-compatible alternatives. We then construct 20 replacement tuples across the three positions. After removing duplicate variants and discarding candidates that modify the harmful payload, we randomly retain eight mutation candidates for each parent template. Each candidate is combined with the unchanged payload and evaluated through a forward pass of the target model. The candidates are then scored by \oracle.

\noindent\textbf{Seed Templates.}
Recent studies in software fuzzing suggest the quality of the initial seed pool can affect fuzzing effectiveness~\cite{herrera2021seed, shen2022drifuzz}. Following previous work~\cite{liu2023jailbreaking}, we collect jailbreak templates from a public repository~\cite{albert_jailbreakchat} and remove duplicate, unsuitable, or payload-modifying templates. After filtering, we retain 64 templates as the initial seed pool. Appendix~\ref{appendix:Jailbreak Seed Templates} provides further details.

\subsection{Datasets and Seed Templates}
\label{subsec:Datasets and Seed Templates}

\noindent\textbf{Target LLMs.}
For the evaluation, we use the 16 recent LLMs from nine providers as shown in Table~\ref{tab:models}. Five models are used as white-box source models, where we construct the \oracle and directly evaluate \ourname under the evaluator setting described in Section~\ref{sec:ThreatModel}. For cross-model transfer, jailbreak templates are optimized on selected source models and then applied zero-shot to target models without target-specific optimization, following the adversary setting. The model set covers both dense and mixture-of-experts (MoE) architectures, as well as Chain-of-Thought (CoT), non-CoT, and hybrid reasoning modes. All transfer models are evaluated with an inference-only setting, whereas proprietary models are accessed through the OpenRouter API~\cite{openrouter2026}.

\noindent\textbf{Oracle Training Data.}
We construct a balanced dataset for activation extraction and \oracle training. The initial dataset contains 7,000 harmful prompts randomly collected from LLM-LAT~\cite{sheshadri2024latent} and sorry-bench~\cite{xie2025sorrybench}, together with 7,000 benign prompts collected from Natural Reasoning~\cite{yuan2025naturalreasoningreasoningwild28m}, OpenAssistant~\cite{kopf2023openassistant}, and Dolly~\cite{DatabricksBlog2023DollyV2}. To reduce template-specific shortcuts as introduced in Section~\ref{subsec:Template-invariant Activation Extraction}, we further construct 1,000 template-invariant pairs using 10 jailbreak templates as described in Appendix~\ref{appendix:Jailbreak Seed Templates}. Each pair contains one harmful payload and one benign payload wrapped by the same template. 
% The final dataset therefore contains 16,000 examples and then used for activation extraction and the selection of safety neurons.

\noindent\textbf{Oracle-ASR Evaluation Data.}
To evaluate the trained \oracle for each target model, we construct two disjoint test sets. The first set contains 4,000 unwrapped prompts (2,000 harmful and 2,000 benign) and measures its classification ability. We use it to evaluate whether \oracle can distinguish harmful from benign inputs. The second set evaluates \oracle under jailbreak templates. It contains 4,000 harmful prompts and a disjoint pool of 20 jailbreak templates. Each harmful prompt is combined with one randomly selected template from this pool. We generate complete target-model responses for these wrapped inputs, measure their response-level attack success rate (ASR), and analyze the relationship between jailbreak success and the corresponding safety alarm scores. The prompts and jailbreak templates in both test sets are disjoint from those used to train \oracle.

\noindent\textbf{Fuzzing Evaluation Data.}
To show the effectiveness of \ourname, we randomly sample 100 harmful questions from the six categories of StrongREJECT~\cite{souly2024strongreject}. These questions are used only for fuzzing and response-level evaluation and do not overlap with the datasets described above.

\subsection{Evaluation Metrics}
\label{subsec:evaluation-metrics}

We use a multi-stage evaluation pipeline to determine whether a generated response constitutes a successful jailbreak. Llama-Guard-4-12B~\cite{grattafiori2024llama} and Qwen3Guard-Gen-8B~\cite{zhao2025qwen3guard} independently evaluate every response. A response is assigned the agreed label when both judges produce the same decision. When their decisions differ, human annotators review the response and determine the final label.
Let $J_{i,j}\in\{0,1\}$ indicate whether template $\tau_j$ successfully jailbreaks the model for harmful payload $p_i$; follow~\cite{yu2024llm}, we define five metrics below.

\begin{table*}[ht]
\centering
\caption{Oracle performance on clean and jailbreak inputs. $\overline{S}$ denotes the mean safety alarm score. $\Delta_{\mathrm{wrap}}$ measures the score reduction after applying
a jailbreak template, and $\Delta_{\mathrm{sep}}$ measures the separation between failed and successful jailbreaks.}
\label{tab:oracleEvaluation}

\small
\setlength{\tabcolsep}{7pt}
\renewcommand{\arraystretch}{0.90}

\begin{tabular}{@{}lrrr|rrrrr@{}}
\toprule
\multirow[c]{2}{*}{\textbf{Target Model}} &
\multicolumn{3}{c|}{\textbf{Clean Classification}} &
\multicolumn{5}{c}{\textbf{Jailbreak Evaluation}} \\
\cmidrule(lr){2-4}
\cmidrule(lr){5-9}
&
\textbf{AUROC} &
$\boldsymbol{\overline{S}_{\mathrm{benign}}}$ &
$\boldsymbol{\overline{S}_{\mathrm{harmful}}}$ &
\textbf{ASR (\%)} &
$\boldsymbol{\Delta_{\mathrm{wrap}}}$ &
$\boldsymbol{\overline{S}_{\mathrm{succ}}}$ &
$\boldsymbol{\overline{S}_{\mathrm{fail}}}$ &
$\boldsymbol{\Delta_{\mathrm{sep}}}$ \\
\midrule

DeepSeek-R1-14B
& 0.997 & 0.001 & 0.999
& 14.20 & 0.116 & 0.557 & 0.937 & 0.380 \\

Llama-3.1-8B-Instruct
& 0.989 & 0.001 & 0.999
& 22.35 & 0.147 & 0.586 & 0.929 & 0.343 \\

Gemma-3-4B-it
& 0.969 & 0.013 & 0.976
& 38.75 & 0.195 & 0.572 & 0.913 & 0.341 \\

Gemma-4-E4B-it
& 0.998 & 0.001 & 0.999
& 1.20 & 0.031 & 0.492 & 0.974 & 0.482 \\

GPT-OSS-20B
& 0.999 & 0.001 & 0.999
& 0.35 & 0.040 & 0.633 & 0.960 & 0.327 \\

\bottomrule
\end{tabular}
\end{table*}

\noindent\textbf{Jailbreak Discovery Rate (JDR).}
JDR measures the ratio of harmful questions for which the fuzzer discovers at least one successful template within the given budget. Let $\mathcal{T}_{B}$ denote all templates discovered within budget $B$. We define
$
\operatorname{JDR}
=
\frac{1}{N}
\sum_{i=1}^{N}
\max_{\tau_j\in\mathcal{T}_{B}} J_{i,j}.
\label{eq:jdr}
$
JDR measures the fuzzer's ability to discover a successful template for each harmful question.

\noindent\textbf{Response Generations per Discovery (RGD).}
RGD measures the average number of target-model responses generated before the first verified jailbreak template. It includes responses generated during successful searches and the final verification. A lower RGD indicates less dependence on autoregressive response generation during the search.

\noindent\textbf{End-to-End Time per Discovery (ETD).}
ETD measures the average time required to discover the first verified jailbreak for a successfully jailbroken question. It includes all components of the fuzzing, such as target-model evaluation, gradient computation, mutation-model inference, \oracle scoring, response generation, and external judge evaluation. We report both RGD and ETD only over successful questions. 

\noindent\textbf{Attack Success Rate (ASR).}
ASR measures the effectiveness of an individual template~\cite{yu2024llm}. Given $N$ harmful questions, we define
$
\operatorname{ASR}(\tau_j)
=
\frac{1}{N}
\sum_{i=1}^{N} J_{i,j}.
\label{eq:asr}
$
A higher ASR indicates that the template succeeds on a larger fraction of the evaluation questions. To evaluate the quality of harmful responses, we also report StrongREJECT rubric~\cite{souly2024strongreject}.

\noindent\textbf{Ensemble Attack Success Rate (EASR).}
EASR measures the combined effectiveness of a small set of templates. Let $\mathcal{T}_{K}$ denote the selected set of $K$ templates. A question is considered successfully tested if at least one template in $\mathcal{T}_{K}$ produces a successful jailbreak. We compute
$
\operatorname{EASR}(\mathcal{T}_{K})
=
\frac{1}{N}
\sum_{i=1}^{N}
\mathbb{I}
\left[
\max_{\tau_j\in\mathcal{T}_{K}} J_{i,j}=1
\right].
\label{eq:easr}
$
EASR represents the percentage of questions that could leverage at least one template from the subset to jailbreak the target LLM.

\subsection{Evaluation Objectives}
\label{subsec:Evaluation Objectives}
We evaluate \ourname{} along four dimensions:

\begin{itemize} [itemsep=0pt, topsep=0pt, leftmargin=*]
    \item \textbf{EO1: Oracle Validity.} Whether the selected safety neurons produce a reliable \oracle{} whose safety alarm score is associated with jailbreak success.
    
    \item \textbf{EO2: Fuzzing Effectiveness and Efficiency.} Whether \oracle{} feedback improves jailbreak discovery while reducing response generation and end-to-end search cost.
    
    \item \textbf{EO3: Universal Template.} Can \ourname optimize a shared template that is effective across harmful payloads?
    
    \item \textbf{EO4: Transferability.} Do optimized templates retain effectiveness on different architectures, scales, reasoning modes, modalities, and datasets without further optimization?
\end{itemize}

\section{Experimental Results}
\label{sec:Experimental Results}

\subsection{EO1: Oracle Validity}
\label{subsec:Oracle Validity}

We evaluate whether the \oracle{} provides a reliable safety signal for fuzzing from three aspects: whether it captures safety-related information, whether this signal remains meaningful under unseen jailbreak contexts, and, most importantly, whether it can rank candidate prompts according to their likelihood of jailbreak success. We first evaluate clean classification on harmful and benign prompts without jailbreak templates. We then evaluate 4,000 harmful prompts wrapped with jailbreak templates. For each wrapped input, we compute its safety alarm score and generate the target-model response to obtain the response-level evaluation result.

\noindent\textbf{Safety Signal Validity.}
Table~\ref{tab:oracleEvaluation} shows that the selected safety neurons provide a strong signal for distinguishing harmful from benign inputs. Across all five source models, the \oracle{} achieves AUROC values between 0.969 and 0.999. Benign prompts receive mean safety alarm scores close to zero, whereas harmful prompts receive scores close to one. This confirms that the learned score captures safety-related information before any jailbreak context is introduced.

\noindent\textbf{Behavior under Jailbreak Contexts.}
We next examine whether the learned signal remains informative when the same harmful payload is embedded in jailbreak templates. The mean safety alarm score decreases on all five models after wrapping, with $\Delta_{\mathrm{wrap}}$ ranging from 0.031 to 0.195. Because the harmful payload itself is unchanged, this consistent decrease suggests that the surrounding jailbreak context weakens the internal signal associated with harmful-intent recognition. This observation is consistent with prior findings that jailbreak contexts can suppress safety-related internal representations~\cite{zhao2026llms,zhou2025role}. More importantly, the alarm score consistently separates successful from failed jailbreak attempts. Although the overall response-level ASR varies substantially across models, from 0.35\% on GPT-OSS-20B to 38.75\% on Gemma-3-4B-it, successful jailbreaks receive lower mean alarm scores than failed attempts on every model. The resulting $\Delta_{\mathrm{sep}}$ ranges from 0.327 to 0.482. For example, on DeepSeek-R1-14B, the mean score decreases from 0.937 for failed attempts to 0.557 for successful jailbreaks. Similar gaps appear across the remaining models. These results indicate that $S_{\mathrm{alarm}}$ captures information related to jailbreak susceptibility beyond harmful-versus-benign classification.

\begin{figure}[t]
  \centering
  \includegraphics[width=0.95\linewidth]{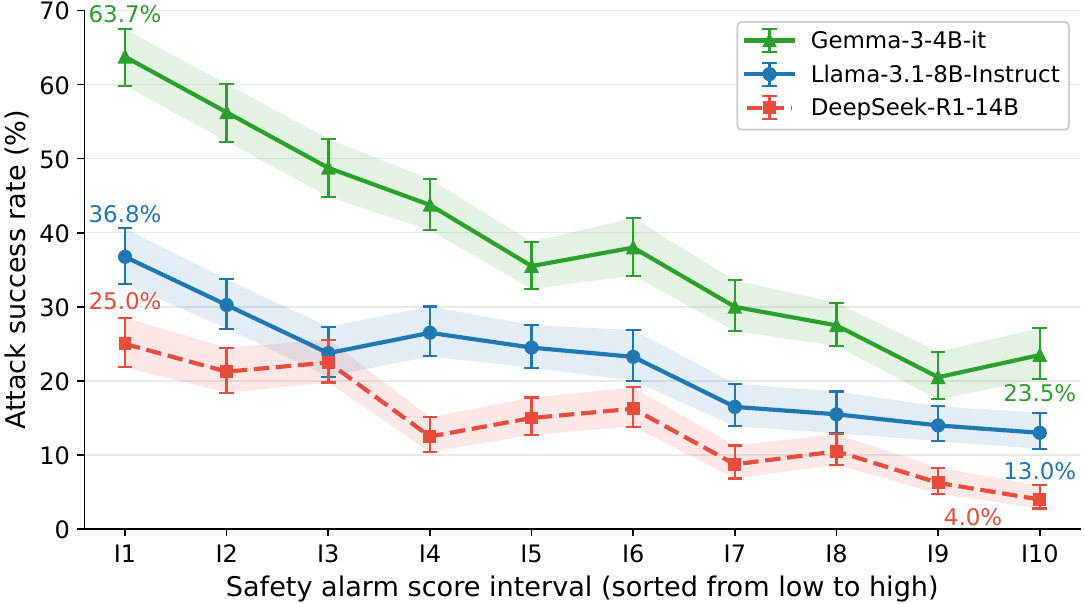}
  \caption{Relationship between the safety alarm score and response-level ASR.}
  \label{img:spearman}
\end{figure}

\noindent\textbf{Correlation with Jailbreak Success.}
For fuzzing, however, separating successful and failed samples on average is not sufficient: the \oracle{} should provide a continuous ranking signal that identifies which candidates are more promising. We therefore further analyze the relationship between $S_{\mathrm{alarm}}$ and response-level ASR. We focus on Llama-3.1-8B-Instruct, DeepSeek-R1-14B, and Gemma-3-4B-it, which provide sufficient successful responses for interval-level analysis. For each model, we sort the 4,000 wrapped inputs by their alarm scores, divide them into ten ordered intervals, and compute the ASR within each interval. Figure~\ref{img:spearman} reports the resulting ASR with 95\% confidence intervals. A clear inverse relationship appears across all three models: intervals with lower safety alarm scores consistently exhibit higher jailbreak success rates. The Spearman coefficients are $-0.964$ for Llama-3.1-8B-Instruct, $-0.927$ for DeepSeek-R1-14B, and $-0.976$ for Gemma-3-4B-it. These strong negative correlations show that $S_{\mathrm{alarm}}$ provides an ordering over candidate jailbreak sensitivity rather than only a binary safety prediction. In other words, even when multiple candidates would receive the same response-level failure label, their alarm scores can still indicate which candidates are closer to successful jailbreaks.

% Together, the results validate $S_{\mathrm{alarm}}$ as continuous feedback for fuzzing. We therefore use the \oracle{} to rank intermediate candidates and guide subsequent mutation \lw{this paragraph can go}.

%%%%%%%%%%%%%%%%%%%%%%%%%%%%%%%%%%%%%%%%%%%%%%%%%%%%%%%%%%
%%%%%%%%%%%%%%%%%%%%%%%%%%%%%%%%%%%%%%%%%%%%%%%%%%%%%%%%%%
%%%%%%%%%%%%%%%%%%%%%%%%%%%%%%%%%%%%%%%%%%%%%%%%%%%%%%%%%%

\subsection{EO2: Fuzzing Effectiveness and Efficiency}
\label{subsec:Fuzzing Effectiveness and Efficiency}

To validate EO2, we evaluate \ourname on the StrongREJECT described in Section~\ref{subsec:Datasets and Seed Templates}. For each harmful payload, we run an independent fuzzing process with a budget of 20 mutation iterations (as shown in Figure~\ref{img:asrScore}, by iteration 20, the ASR has stabilized and safety alarm score decreases slowly). \ourname evaluates intermediate candidates using the \oracle and generates a target-model response only for final verification. The search stops when the budget is exhausted. We report the seed JDR of all $64 \times 100$ samples. We compare \ourname with four representative automated jailbreak methods, GCG~\cite{zou2023universal}, AutoDAN~\cite{liu2024autodan}, PAIR~\cite{chao2025jailbreaking}, and LLM-Fuzzer~\cite{yu2024llm}. These methods cover token-level gradient optimization, genetic prompt search, attacker-LLM refinement, and fuzzing-based template mutation. Appendix~\ref{appendix:Baseline Methods Setup} provides the complete configurations of baseline methods.

\begin{table}[!t]
\centering
\caption{Effectiveness and efficiency of \ourname on different source models. Best results are shown in \textbf{bold}, and second-best results are \underline{underlined}.}
\label{tab:sourceResults}

\small
\setlength{\tabcolsep}{5pt}
\renewcommand{\arraystretch}{0.95}

\begin{tabular}{@{}l|l|rrr@{}}
\toprule
\textbf{Target Model} &
\textbf{Method} &
\textbf{JDR $\uparrow$} &
\textbf{RGD $\downarrow$} &
\textbf{ETD (s) $\downarrow$} \\
\midrule

\multirow{6}{*}{DeepSeek-R1-14B}
& Seed & 18\%  & --            & -- \\
& GCG           & 19\%  & \textbf{1.0}  & 587.2 \\
& AutoDAN       & 47\%  & \underline{14.7} & 470.3 \\
& PAIR          & \underline{94\%} & 15.6 & \textbf{43.4} \\
& LLM-Fuzzer    & 93\%  & 179.6         & 114.5 \\
& \ourname{}    & \textbf{100\%} & \textbf{1.0} & \underline{64.1} \\
\midrule

\multirow{6}{*}{\shortstack[l]{Llama-3.1-\\8B-Instruct}}
& Seed & 25\%  & --            & -- \\
& GCG           & 91\%  & \textbf{1.0}  & 621.7 \\
& AutoDAN       & 86\%  & 31.8          & 557.2 \\
& PAIR          & 81\%  & \underline{17.2} & \underline{51.4} \\
& LLM-Fuzzer    & \underline{97\%} & 214.7 & 97.7 \\
& \ourname{}    & \textbf{100\%} & \textbf{1.0} & \textbf{50.7} \\
\midrule

\multirow{6}{*}{Gemma-3-4B-it}
& Seed & 36\%  & --            & -- \\
& GCG           & \underline{97\%} & \textbf{1.0} & 634.1 \\
& AutoDAN       & 94\%  & 22.3          & 541.6 \\
& PAIR          & \textbf{100\%} & \underline{17.8} & \textbf{53.7} \\
& LLM-Fuzzer    & \textbf{100\%} & 183.9 & 106.8 \\
& \ourname{}    & \textbf{100\%} & \textbf{1.0} & \underline{61.8} \\
\midrule

\multirow{6}{*}{Gemma-4-E4B-it}
& Seed & 4\%   & --            & -- \\
& GCG           & 28\%  & \textbf{1.0}  & 612.9 \\
& AutoDAN       & 21\%  & 53.8          & 635.1 \\
& PAIR          & 37\%  & \underline{22.3} & \underline{58.5} \\
& LLM-Fuzzer    & \underline{49\%} & 243.6 & 124.7 \\
& \ourname{}    & \textbf{76\%} & \textbf{1.0} & \textbf{56.3} \\
\midrule

\multirow{6}{*}{GPT-OSS-20B}
& Seed & 0\%   & --            & -- \\
& GCG           & 0\%   & --            & -- \\
& AutoDAN       & 19\%  & 45.2          & 725.2 \\
& PAIR          & 43\%  & \underline{18.6} & \textbf{44.7} \\
& LLM-Fuzzer    & \underline{48\%} & 304.1 & 133.6 \\
& \ourname{}    & \textbf{96\%} & \textbf{1.0} & \underline{72.5} \\
\bottomrule
\end{tabular}

\vspace{3pt}
\begin{minipage}{1\columnwidth}

\end{minipage}
\end{table}

As shown in Table~\ref{tab:sourceResults}, \ourname achieves the highest JDR on all five source models. It reaches 100\% JDR on Llama-3.1-8B-Instruct, DeepSeek-R1-14B, and Gemma-3-4B-it. On the two more strongly aligned models, it achieves 76\% JDR on Gemma-4-E4B-it and 96\% on GPT-OSS-20B, exceeding the evaluated methods by 27 and 48 percentage points, respectively. On Gemma-3-4B-it, \ourname, PAIR, and LLM-Fuzzer all achieve 100\% JDR, while \ourname requires fewer response generations. The advantage is particularly clear on Gemma-4-E4B-it and GPT-OSS-20B, where the initial seed templates achieve only 4\% and 0\% JDR, respectively. Response-level fuzzers rely on discrete success or failure feedback from generated responses. When successful generations are rare, this binary signal provides little information for distinguishing promising failed candidates, which limits the effectiveness of subsequent seed selection and mutation. GCG and AutoDAN avoid response-level feedback by optimizing the likelihood of affirmative response prefixes. However, this surrogate objective has become less reliable on recent aligned and reasoning models, which may produce affirmative prefixes or extended reasoning without ultimately generating genuinely harmful content~\cite{qi2025safety,zhu2026advprefix,li2025safety,xu2026reasoning}. In contrast, the \oracle{} assigns a continuous score to every candidate before response generation. It therefore provides informative feedback even when none of the current candidates produces a successful jailbreak, helping \ourname maintain high discovery coverage on these more aligned models.

\ourname maintains an RGD of 1.0 as it evaluates intermediate candidates through prefill-only \oracle{} scoring and generates the response only for final verification. GCG also achieves an RGD of 1.0 when successful, but its suffix-level optimization requires substantially more computation and performs poorly on the more strongly aligned models. AutoDAN requires between 14.7 and 53.8 responses per discovery, PAIR requires between 15.6 and 22.3, and LLM-Fuzzer requires between 179.6 and 304.1. These results show that the \oracle{} removes the need to repeatedly generate target-model responses as mutation feedback. 

This reduction in response generation also leads to end-to-end time efficiency. \ourname{} achieves the lowest average ETD on Llama-3.1-8B-Instruct and Gemma-4-E4B-it, and the second-lowest ETD on the other three models. PAIR has a lower conditional ETD on successfully jailbroken questions for DeepSeek-R1-14B and GPT-OSS-20B, but its JDR is lower by 6 and 53 percentage points, respectively. GCG and AutoDAN require several hundred seconds per discovery because they repeatedly evaluate large token-level or population-level candidate sets. \ourname{} additionally incurs a small one-time offline cost to construct the \oracle{}. As shown in Appendix~\ref{appendix:Time Consumption of Oracle}, this process takes only 18 minutes and 32 seconds on average and can be further used in subsequent testing. Overall, \ourname{} improves jailbreak discovery coverage, greatly reduces complete response generation, and maintains low end-to-end search cost with only a modest one-time preparation overhead.

\subsection{EO3: Universal Template Optimization}
\label{subsec:Universal Template Optimization}

\begin{figure}[ht]
  \centering
  \includegraphics[width=0.95\linewidth]{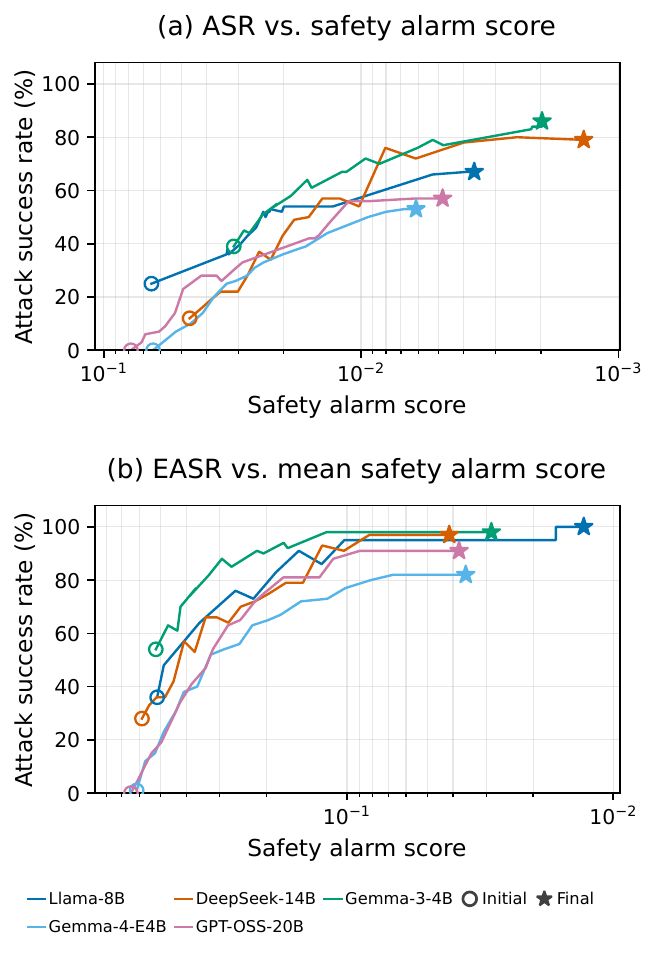}
  \caption{ASR and EASR versus the safety alarm score during 20 fuzzing steps.}
  \label{img:asrScore}
\end{figure}
Section~\ref{subsec:Fuzzing Effectiveness and Efficiency} optimizes a separate jailbreak template for each harmful payload. We next evaluate whether \ourname{} can optimize universal templates that generalize across payloads. For each source model, we optimize shared templates over a disjoint set of 100 harmful payloads~\cite{souly2024strongreject} for 20 iterations, using the mean \oracle{} safety alarm score across payloads as the objective. Following prior work~\cite{yu2024llm}, we report ASR for the lowest-scoring template and EASR for the five lowest-scoring templates. This setting evaluates whether \ourname{} can discover a small set of reusable jailbreak templates.

Figure~\ref{img:asrScore} shows the optimization trajectories. Across all models, decreases in the mean safety alarm score are accompanied by consistent improvements in ASR and EASR, showing that the \oracle{} provides an effective optimization signal for template search. The EASR reaches at least 90\% on four models and 82\% on Gemma-4-E4B-it, indicating that a small set of optimized templates can cover a large fraction of harmful payloads. Most gains are achieved within the early iterations, and both attack performance and the safety alarm score become relatively stable near the end of the 20-iteration budget. Overall, these results show that \ourname{} can efficiently optimize reusable jailbreak templates that generalize across different harmful payloads on the same source model.

\begin{table}[t]
\centering
\caption{StrongREJECT rubric evaluation across models.}
\label{tab:strongrejectResults}
\small
\setlength{\tabcolsep}{7pt}
\renewcommand{\arraystretch}{0.95}

\begin{tabular}{lccc}
\toprule
\textbf{Model} &
\textbf{StrongREJECT} &
\textbf{Conv.} &
\textbf{Spec.} \\
\midrule

DeepSeek-R1-14B
& 0.647 & 4.32 & 4.23 \\

Llama-3.1-8B-Instruct
& 0.559 & 4.42 & 4.26 \\

Gemma-3-4B-it
& 0.714 & 4.28 & 4.36 \\

Gemma-4-E4B-it
& 0.425 & 4.24 & 4.18 \\

GPT-OSS-20B
& 0.445 & 4.11 & 4.13 \\

\midrule
\textit{Average}
& 0.558 & 4.27 & 4.23 \\

\bottomrule
\end{tabular}
\end{table}

\noindent\textbf{StrongREJECT Rubric.}
Prior work has shown that some jailbreaks achieve non-refusal without eliciting substantive harmful information~\cite{souly2024strongreject}. We further evaluate the quality of successful responses using StrongREJECT rubric, which measures convincingness and specificity in addition to refusal. Appendix~\ref{appendix:StrongREJECT Rubric} provides the complete scoring setup. As shown in Table~\ref{tab:strongrejectResults}, the average StrongREJECT score is 0.558, with convincingness and specificity scores of 4.27 and 4.23 out of 5, respectively. These results indicate that the templates discovered by \ourname elicit substantive harmful responses rather than merely bypassing refusal.

\subsection{EO4: Cross-Model Transferability}
\label{subsec:Cross-Model Transferability}

To answer EO4, we jointly optimize jailbreak templates on Llama-3.1-8B-Instruct and GPT-OSS-20B (Appendix~\ref{appendix:Source Model Selection for Transfer Optimization} justifies this choice). We follow the fuzzing procedure described in Section~\ref{subsec:Universal Template Optimization} and run the search for 160 iterations (where the safety alarm scores have converged). During each iteration, both source-model oracles evaluate each candidate, and we use their mean safety alarm score as feedback for seed selection and mutation. After optimization, we directly apply the resulting templates to the evaluation models without further mutation or model-specific tuning. We repeat each evaluation five times with a generation temperature of 1.0 and report the mean and standard deviation of top-1 ASR and top-5 EASR.

% Place this command in the preamble.
\newcommand{\result}[2]{%
  \ensuremath{#1\%_{\scriptscriptstyle \pm #2\%}}%
}

\begin{table}[ht]
\centering
\caption{Universal-template transfer evaluation across target models over five runs.}
\label{tab:transferResults}
\small
\setlength{\tabcolsep}{7pt}
\renewcommand{\arraystretch}{0.95}

\begin{adjustbox}{max width=\columnwidth}
\begin{tabular}{l|c|c|c}
\toprule
\textbf{Target Model} &
\textbf{Seeds} &
\textbf{ASR} &
\textbf{EASR} \\
\midrule

DeepSeek-R1-14B
& $4\%$
& \result{79.0}{1.9}
& \result{98.6}{1.1} \\

Gemma-3-4B-it
& $8\%$
& \result{85.8}{2.3}
& \result{98.2}{1.3} \\

Gemma-4-E4B-it
& $0\%$
& \result{43.8}{2.2}
& \result{70.4}{3.0} \\

Llama-3.2-3B-Instruct
& $2\%$
& \result{76.0}{3.5}
& \result{98.6}{1.7} \\

Phi-4
& $0\%$
& \result{70.8}{2.5}
& \result{91.4}{3.5} \\

Qwen3.6-35B-A3B
& $0\%$
& \result{67.6}{3.5}
& \result{98.0}{1.6} \\

Qwen3.6-27B
& $0\%$
& \result{63.2}{2.3}
& \result{90.2}{3.6} \\

GLM-4.7-Flash
& $0\%$
& \result{70.4}{3.4}
& \result{95.0}{1.6} \\

\midrule

\textit{Inference Average}
& -- 
& \textbf{69.6\%}
& \textbf{92.6\%} \\

\midrule

LongCat-2.0
& $0\%$
& \result{66.8}{4.1}
& \result{95.6}{2.9} \\

DeepSeek-V4-Flash
& $0\%$
& \result{53.0}{2.0}
& \result{71.2}{1.5} \\

DeepSeek-V4-Pro
& $0\%$
& \result{17.6}{2.4}
& \result{28.8}{1.5} \\

Grok-4.5
& $0\%$
& \result{17.2}{0.4}
& \result{28.6}{2.3} \\

Gemini-2.5-Pro
& $0\%$
& \result{85.2}{1.9}
& \result{98.2}{2.2} \\

Gemini-3-Flash
& $0\%$
& \result{24.6}{1.5}
& \result{37.4}{1.3} \\

\midrule
\textit{Proprietary Average}
& -- 
& \textbf{44.1\%}
& \textbf{60.0\%} \\

\bottomrule
\end{tabular}
\end{adjustbox}
\end{table}

Table~\ref{tab:transferResults} reports the cross-model transfer results. Harmful queries without jailbreak templates achieve at most 8\% ASR and fail completely on 11 of the 14 models. In contrast, the optimized templates achieve transfer without any target-specific optimization. Transfer is particularly strong across the eight inference-only target models. The optimized top-1 template achieves an average ASR of 69.6\%, while the top-5 template set reaches an average EASR of 92.6\%. Seven of the eight models achieve EASR above 90\%, including 98.6\% on Llama-3.2-3B-Instruct and 98.0\% on Qwen3.6-35B-A3B. Strong transfer is observed across different model families, parameter scales, reasoning modes, and both dense and MoE architectures. These results show that the optimized templates transfer well beyond the source models and remain effective across different target LLMs.

The optimized templates also transfer to proprietary models. Across the six API models, the average ASR and EASR are 44.1\% and 60.0\%, respectively. Gemini-2.5-Pro shows the strongest transfer, reaching 85.2\% ASR and 98.2\% EASR. The templates also achieve 66.8\% ASR and 95.6\% EASR on LongCat-2.0 and 53.0\% ASR and 71.2\% EASR on DeepSeek-V4-Flash. Transfer is weaker on DeepSeek-V4-Pro, Grok-4.5, and Gemini-3-Flash, but all three models still exhibit non-zero ASR and EASR despite receiving no target-specific optimization. Overall, these results show that templates optimized using white-box surrogate models can retain meaningful effectiveness when transferred directly to proprietary models accessible only through their input-output interfaces. We also evaluate cross-dataset transfer on HarmBench in Appendix~\ref{appendix:Cross-Dataset Evaluation on HarmBench}, which retains 67.0\% ASR and 87.3\% EASR.

\begin{table}[ht]
\centering
\caption{Transferability of \ourname on VLMs.}
\label{tab:vlm}
\small
\setlength{\tabcolsep}{4pt}
\renewcommand{\arraystretch}{0.95}

\begin{adjustbox}{max width=\columnwidth}
\begin{tabular}{@{}l|cc|cc@{}}
\toprule
\multirow{2}{*}{\textbf{Target Model}} &
\multicolumn{2}{c|}{\textbf{Baseline}} &
\multicolumn{2}{c}{\textbf{\ourname} ASR / EASR} \\
\cmidrule(lr){2-3}
\cmidrule(lr){4-5}
&
\textbf{T2I} &
\textbf{NSFW} &
\textbf{T2I} &
\textbf{NSFW} \\
\midrule

Qwen2.5-VL-3B-Instruct
& 2\% & 6\% & 73\% / 93\% & 88\% / 100\% \\

Qwen2.5-VL-7B-Instruct
& 0\% & 4\% & 65\% / 81\% & 63\% / 89\% \\

Qwen3.5-35B-A3B
& 1\% & 0\% & 41\% / 72\% & 69\% / 82\% \\

Kimi-VL-A3B-Instruct
& 5\% & 8\% & 86\% / 100\% & 83\% / 100\% \\

Gemma-4-31B-it
& 0\% & 0\% & 57\% / 83\% & 71\% / 94\% \\

\midrule
\textit{Average}
& 1.6\%
& 3.6\%
& 64.4\% / 85.8\%
& 74.8\% / 93.0\% \\

\bottomrule
\end{tabular}
\end{adjustbox}
\end{table}

\noindent\textbf{Transferability on Different Modalities.} We further evaluate whether the templates optimized by \ourname transfer to multimodal safety tasks. We consider five Visual Language Models (VLMs): Qwen2.5-VL-3B-Instruct and Qwen2.5-VL-7B-Instruct~\cite{wu2025qwen}, Qwen3.5-35B-A3B~\cite{qwen3.5}, Kimi-VL-A3B-Instruct~\cite{team2025kimi}, and Gemma-4-31B-it~\cite{gemma4technicalreport}. We reuse the templates optimized in the previous text-only fuzzing without further adaptation. We evaluate two visual tasks. For text-to-image (T2I), we randomly sample 100 harmful instructions from HarmBench~\cite{mazeika2024harmbench} and convert each instruction into an image following prior works~\cite{gong2025figstep, liu2025survey}. The accompanying text prompt is ``Answer the question based on the image.'' For NSFW evaluation, we sample 100 unsafe images from the NSFW-1024 dataset~\cite{yesidobyte_nsfw1024} and use ``Describe the image.'' as the payload. The baseline uses the same images and payloads without a jailbreak template. As shown in Table~\ref{tab:vlm}, the baseline ASR is low across all models, averaging only 1.6\% on T2I and 3.6\% on NSFW. After adding the transferred templates, the average ASR increases to 64.4\% on T2I and 74.8\% on NSFW, while the corresponding EASR reaches 85.8\% and 93.0\%, respectively. These results show that templates optimized from text-only internal safety signals can transfer to visual tasks without access to the target VLM's activations. The higher performance on NSFW images further suggests that the optimized templates generalize beyond textual harmful instructions to image-specific unsafe content.

\section{Ablation Study}
\label{sec:Ablation and Hyperparameter Study}

%We further study the sensitivity of the method to the number of selected neurons and the fuzzing budget.

\subsection{Safety-Neuron Construction}
\label{ablation:Neuron Selection}

We examine the two designs used to construct the safety-neuron set: template-invariant activation extraction (Section~\ref{subsec:Template-invariant Activation Extraction}) and stability-aware neuron selection (Section~\ref{subsec:Stability-Aware Safety Neuron Selection}). We conduct the ablation on Llama-3.1-8B-Instruct while keeping other configurations fixed. \textit{Prompt-pair only} extracts activations from harmful and benign prompts without introducing jailbreak-template variations. \textit{Z-score} replaces stability selection with statistic-based ranking method~\cite{wu2025neurostrike}. \textit{Random} selects the same number of neurons uniformly at random from the corresponding layers. We report clean-prompt AUROC, the Spearman correlation between the safety alarm score and ASR, and the downstream JDR.

\begin{table}[ht]
\centering
\caption{Ablation of safety-neuron extraction and selection.}
\label{tab:ablationNeurons}
\small
\setlength{\tabcolsep}{3pt}
\renewcommand{\arraystretch}{0.95}

\resizebox{\columnwidth}{!}{
\begin{tabular}{@{}llccc@{}}
\toprule
\textbf{Extraction} &
\textbf{Selection} &
\textbf{AUROC $\uparrow$} &
$\boldsymbol{\rho \downarrow}$ &
\textbf{JDR $\uparrow$} \\
\midrule

Prompt-pair only
& Stability
& 0.834 & -0.437 & 74\% \\

Template-invariant
& Z-score
& 0.974 & -0.737 & 82\% \\

Template-invariant
& Random
& 0.682 & 0.121 & 16\% \\

\textbf{Template-invariant}
& \textbf{Stability}
& \textbf{0.989} & \textbf{-0.964} & \textbf{100\%} \\

\bottomrule
\end{tabular}
}
\end{table}

Table~\ref{tab:ablationNeurons} shows that the full \ourname{} achieves the best performance across all three metrics. Removing template-invariant extraction reduces AUROC from 0.989 to 0.834, weakens the correlation from $-0.964$ to $-0.437$, and decreases JDR from 100\% to 74\%. This indicates that neurons identified only from clean prompts do not generalize well to harmful inputs under jailbreak contexts, while including template variations helps preserve informative safety signals across different prompt contexts. Replacing stability selection with Z-score ranking retains a high clean-prompt AUROC of 0.974, but reduces the correlation to $-0.737$ and JDR to 82\%. This suggests that classification accuracy alone does not guarantee a useful fuzzing signal. Stability-aware selection favors neurons whose associations remain consistent across training samples, producing a score that better ranks jailbreak candidates. Random neuron selection performs substantially worse, with an AUROC of 0.682, a positive correlation of 0.121, and only 16\% JDR. Overall, template-invariant extraction improves robustness to jailbreak contexts, while stability selection provides a more reliable signal for fuzzing.

\subsection{Oracle-Guided Fuzzing}
\label{ablation:SafetyOracle}

We examine the roles of continuous oracle feedback and gradient-guided mutation on well-aligned GPT-OSS-20B. The \textit{w/o gradient guidance} variant retains continuous \oracle scoring but selects mutation positions randomly; the \textit{binary feedback} variant replaces \oracle with response-level success labels and follows the ranking strategy of LLM-Fuzzer. Since binary feedback is non-differentiable, it also uses random token positions. All variants use the same seeds, mutation model, and candidate budget. We use 100 iterations instead of the default 20 to provide the weakened variants with sufficient opportunity to discover successful mutations.

\begin{table}[ht]
\centering
\caption{Ablation of continuous oracle feedback and gradient-guided mutation on GPT-OSS-20B with 100 iterations.}
\label{tab:ablationOracle}
\small
\setlength{\tabcolsep}{3pt}
\renewcommand{\arraystretch}{0.95}

\resizebox{\columnwidth}{!}{
\begin{tabular}{lccccc}
\toprule
\textbf{Variant} &
\textbf{\shortstack{Search\\Feedback}} &
\textbf{\shortstack{Token\\Position}} &
% \textbf{\shortstack{Candidate\\Ranking}} &
\textbf{JDR $\uparrow$} &
\textbf{RGD $\downarrow$} \\
\midrule

W/o gradient guidance
& Continuous
& Random
% & Lowest score
& 21\% & 1.0 \\

Binary feedback
& Binary
& Random
% & LLM-Fuzzer~\cite{yu2024llm}
& 4\% & 18.3 \\

Full \ourname
& Continuous
& Gradient
% & Lowest score
& 99\% & 1.0 \\

\bottomrule
\end{tabular}
}
\end{table}

As shown in Table~\ref{tab:ablationOracle}, the full method reaches 99\% JDR, while removing gradient guidance reduces JDR to 21\%. Both variants maintain an RGD of 1.0 because intermediate candidates are evaluated through prefill-only oracle scoring. This gap shows that the continuous alarm score supports not only candidate ranking but also sensitive-token localization through its gradients. Without a differentiable score, token-level sensitivity cannot be estimated. The random-position variant still achieves 21\% JDR, reflecting the sensitivity of LLMs to small input changes~\cite{robey2023smoothllm,salinas2024butterfly} where random substitutions occasionally create unusual local patterns or shift the prompt away from the original template distribution, thereby weakening refusal behavior~\cite{andriushchenko2025jailbreaking}. However, such useful mutations are infrequent, whereas gradient guidance systematically targets positions that most strongly affect the internal safety signal. Replacing the continuous oracle with binary feedback further reduces JDR to 4\% and increases RGD to 18.3. Because successful responses are near-zero on GPT-OSS-20B, most candidates receive the same failure label, providing little information for mutation. In contrast, the SafetyOracle assigns a continuous score before decoding, allowing \ourname to distinguish and retain promising candidates even before a successful jailbreak is observed.

\section{Defenses}
\label{sec:Defenses}

\subsection{Perplexity Filter \& SmoothLLM}
\label{subsec:Perplexity Filter SmoothLLM}
A perplexity filter rejects inputs that are unlikely under a reference language model. Following~\cite{alon2023detecting}, we use GPT-2 and reject prompts whose perplexity exceeds the selected threshold. SmoothLLM generates perturbed copies of each prompt and aggregates their response-level decisions~\cite{robey2023smoothllm}. We use the Swap perturbation with a perturbation ratio of 10\% and generate 10 copies for each prompt.

\begin{table}[ht]
\centering
\caption{\ourname under input defenses.}
\label{tab:defenseResults}
\small
\setlength{\tabcolsep}{2.8pt}
\renewcommand{\arraystretch}{0.95}

\begin{tabular}{lccc}
\toprule
\textbf{Model} &
\textbf{No Defense} &
\textbf{Perplexity} &
\textbf{SmoothLLM} \\
\midrule

DeepSeek-R1-14B
& 100\%
& \dropval{98}{2}
& \dropval{74}{26} \\

Llama-3.1-8B-Instruct
& 100\%
& \dropval{95}{5}
& \dropval{71}{29} \\

Gemma-3-4B-it
& 100\%
& \dropval{92}{8}
& \dropval{75}{25} \\

Gemma-4-E4B-it
& 76\%
& \dropval{72}{4}
& \dropval{65}{11} \\

GPT-OSS-20B
& 96\%
& \dropval{91}{5}
& \dropval{79}{17} \\

\bottomrule
\end{tabular}

\vspace{2pt}
\begin{minipage}{0.98\columnwidth}
\end{minipage}
\end{table}

As shown in Table~\ref{tab:defenseResults}, the perplexity filter causes only a small reduction in the effectiveness of \ourname. Across the five models, JDR decreases by only 2--8\%. This robustness arises because \ourname uses the masked language model to replace selected tokens with context-compatible alternatives while preserving the length and grammatical structure of the original template. Consequently, the optimized templates remain fluent and do not rely on anomalous high-perplexity sequences~\cite{zou2023universal}. SmoothLLM provides a stronger defense, reducing JDR by 11--29\%. Character-level swaps can disrupt important words and local structures within the optimized templates, making this defense more effective than perplexity filtering. Nevertheless, \ourname retains JDRs of 71\%, 74\%, 75\%, 65\%, and 79\% across the five models. Since \ourname modifies only a small number of sensitive token positions, randomly selected character perturbations may not consistently disrupt the key structures responsible for the jailbreak. Overall, the results show that input perturbation reduces attack effectiveness but does not fully eliminate the vulnerabilities exposed by \ourname. We also compare with other methods under the input defenses in Appendix~\ref{appendix:Robustness of Baseline Attacks to Input Defenses} where \ourname achieves the highest JDR in both settings.

\subsection{Input-Side Jailbreak Detection}
\label{subsec:jailbreak-detection}

To further show the use of \oracle, we use it as an input-side detector for jailbreak prompts. The results in Section~\ref{subsec:Oracle Validity} reveal three distinct alarm-score patterns. Benign prompts usually receive scores close to zero, while direct harmful prompts receive scores close to one. Successful jailbreak prompts often fall between these two regions, suggesting that the surrounding jailbreak template weakens, but does not fully remove, the model's internal recognition of harmful intent. Based on this observation, we define the jailbreak detection score as
$
D_{\mathrm{jb}}(x)
=
1-\left|2S_{\mathrm{alarm}}(x)-1\right|
\in [0,1].
$
The score reaches its maximum when $S_{\mathrm{alarm}}(x)=0.5$ and decreases as the alarm score approaches either zero or one. Therefore, a higher $D_{\mathrm{jb}}(x)$ indicates that the input is closer to the intermediate alarm-score region associated with successful jailbreaks. We evaluate the detector using the 4,000 test jailbreak inputs and 4,000 benign and harmful requests described in Section~\ref{subsec:Datasets and Seed Templates}. We use the response-level evaluation pipeline to assign the ground-truth label, treating successful jailbreaks as positive samples, where normal benign requests and failed jailbreak attempts as negative samples.

\begin{table}[ht]
\centering
\caption{Performance of \oracle{} for detecting successful jailbreak prompts. TPR is reported at a fixed FPR of 5\%.}
\label{tab:jailbreakDetection}
\small
\setlength{\tabcolsep}{5pt}
\renewcommand{\arraystretch}{0.95}

\begin{adjustbox}{max width=\columnwidth}
\begin{tabular}{lcccc}
\toprule
\textbf{Target Model} &
\textbf{AUROC} &
\textbf{Precision} &
\textbf{F1} &
\textbf{TPR@5\%FPR} \\
\midrule

Llama-3.1-8B-Instruct
& 0.957 & 0.884 & 0.892 & 88.4\% \\

DeepSeek-R1-14B
& 0.948 & 0.867 & 0.876 & 86.1\% \\

Gemma-3-4B-it
& 0.963 & 0.896 & 0.901 & 89.3\% \\

\midrule
\textit{Average}
& 0.956 & 0.882 & 0.890 & 87.9\% \\

\bottomrule
\end{tabular}
\end{adjustbox}
\end{table}

As shown in Table~\ref{tab:jailbreakDetection}, the detector achieves AUROC values between 0.948 and 0.963 across the three models. Its average precision and F1 score are 0.882 and 0.890, respectively. At a fixed false-positive rate of 5\%, it detects 87.9\% of successful jailbreak prompts on average. These results show that the \oracle provides a useful input-side signal for identifying likely successful jailbreaks before response generation. Unlike response-level judges, which require complete target-model outputs, the \oracle detector uses only a prefill pass. Therefore, \oracle can serve as a lightweight and convenient detector that provides useful signals for subsequent verification.

\section{Conclusion}
\label{sec:Conclusion}

We presented \ourname{}, a white-box fuzzing framework that uses internal neuron activations as continuous feedback for LLM safety evaluation. By constructing a lightweight \oracle{} and using its differentiable safety alarm score to guide candidate ranking and targeted template mutation, \ourname{} reduces reliance on repeated response generation while preserving the harmful payload and overall prompt semantics with strong transferability. Our evaluation shows that internal safety representations provide informative signals for jailbreak discovery and support effective generalization across different payloads, models, and modalities. Beyond fuzzing, the \oracle{} can also serve as an input-side detector for identifying likely jailbreak prompts before response generation. Overall, our findings demonstrate that internal model signals can serve as practical execution feedback for LLM fuzzing while also providing useful defensive signals for systematic safety evaluation.

\appendix

\bibliographystyle{plainurl}
\bibliography{references}

\section{Jailbreak Seed Templates}
\label{appendix:Jailbreak Seed Templates}

As described in Section~\ref{subsec:Datasets and Seed Templates}, we followed the sampling method used in previous work~\cite{liu2023jailbreaking}. We collected jailbreak templates from a shared online repository~\cite{albert_jailbreakchat}, which contained 79 valid templates. However, several templates were highly similar, especially variants derived from the DAN prompt. We therefore manually removed duplicate or near-duplicate templates. We also excluded templates that directly modified the input payload, as such modifications could introduce additional variables and reduce the fairness of the evaluation. After the filtering process, we retained 74 jailbreak templates, 10 of which were held out for oracle training and the rest 64 templates were used for further fuzzing evaluation. Table~\ref{tab:jailbreakTemplate} shows the categories of templates.

\begin{table}[htbp]
    \centering
    \caption{Distribution of the selected templates by category.}
    \setlength{\tabcolsep}{10pt}
    \renewcommand{\arraystretch}{0.95}
    \label{tab:jailbreakTemplate}
    \begin{tabular}{l c}
        \toprule
        \textbf{Category} & \textbf{Number of Templates} \\
        \midrule
        Role-playing & 33 \\
        Fictional scenarios & 7 \\
        Instruction conflicts & 14 \\
        Refusal suppression & 4 \\
        Emotional framing & 6 \\
        \midrule
        \textit{Total} & 64 \\
        \bottomrule
    \end{tabular}
\end{table}

\section{Baseline Methods}
\label{appendix:Baseline Methods Setup}

We implement four representative jailbreak baselines: GCG~\cite{zou2023universal}, AutoDAN~\cite{liu2024autodan}, PAIR~\cite{chao2025jailbreaking}, and LLM-Fuzzer~\cite{yu2024llm}. We use their released implementations and preserve their main optimization procedures. All methods are evaluated on the same harmful payloads, hardware, and final response-level evaluation pipeline as \ourname{}.

\noindent\textbf{GCG.}
We implement GCG following the original work and its official implementation~\cite{zou2023universal}. GCG appends an optimizable adversarial suffix to each harmful payload and minimizes the negative log-likelihood of a predefined affirmative completion (``Sure, here is''). We initialize the suffix with 20 repeated exclamation-mark tokens and optimize it for 500 steps. At each step, token gradients identify the top 256 replacement tokens, from which a batch of 512 candidate substitutions is evaluated. The candidate with the lowest target loss is retained for the next step. During evaluation, the optimized suffix is directly appended to the harmful payload.

\noindent\textbf{AutoDAN.}
We use the hierarchical genetic algorithm version of AutoDAN~\cite{liu2024autodan}. AutoDAN initializes a population from handcrafted jailbreak prompts and evolves them through hierarchical crossover and mutation. Candidates are ranked by the negative log-likelihood of an affirmative target response. Following the original configuration, we use a population size of 256, a crossover rate of 0.5, a mutation rate of 0.01, an elite rate of 0.1, and five crossover points. We run the optimization for 100 iterations, corresponding to up to 25,600 population-level candidate evaluations. The candidate with the lowest optimization loss is retained for final evaluation.

\noindent\textbf{PAIR.}
We implement PAIR using its released framework and original search configuration~\cite{chao2025jailbreaking}. PAIR uses an attacker LLM to iteratively refine a jailbreak prompt based on the target model's response and evaluation feedback. We use Mixtral-8x7B-Instruct as the attacker model with a temperature of 1.0 and top-$p$ of 0.9. We run 30 parallel conversation streams with a maximum depth of three iterations, resulting in at most 90 target-model queries per payload. The search stops early when a successful jailbreak is identified or when all streams reach the maximum depth.

\noindent\textbf{LLM-Fuzzer.}
We implement LLM-Fuzzer using its released framework and default MCTS-Explore seed-selection strategy~\cite{yu2024llm}. For a fair comparison, we initialize it with the same 64 jailbreak templates used by \ourname{}. At each iteration, one seed is selected and mutated using one of five LLM-based operators: generate, crossover, expand, shorten, or rephrase. We use gpt-5.4-mini-2026-03-17~\cite{openai2026gpt54thinking} as the mutation model. Each generated template is combined with the same evaluation payloads, and the target-model responses are evaluated to update the MCTS-Explore tree. We run LLM-Fuzzer for 500 mutation iterations. 

\section{Computation Statement}
\label{appendix:Time Consumption of Oracle}
All experiments were conducted on a high-performance computing cluster equipped with four NVIDIA GH200 nodes. Each node provides one NVIDIA H100 GPU with 120\,GB of memory. Most experiments were run on a single node, while large-scale evaluations were distributed across multiple nodes to reduce the overall runtime. For black-box API evaluation, all models were accessed through OpenRouter\footnote{\url{https://openrouter.ai/}}. All locally deployed models were obtained from Hugging Face\footnote{\url{https://huggingface.co/}}and evaluated using their unquantized model weights.

\subsection{SafetyOracle Construction Cost}

Table~\ref{tab:oracleConstructionCost} reports the one-time offline cost of constructing the \oracle{} for each source model. The process includes activation extraction, stability-aware neuron selection, and fitting the final Elastic Net classifier. All measurements report wall-clock time on a single NVIDIA GH200 node.

\begin{table}[ht]
\centering
\caption{One-time offline cost of constructing the \oracle{} for each source model.}
\label{tab:oracleConstructionCost}
\scriptsize
\setlength{\tabcolsep}{3pt}
\renewcommand{\arraystretch}{0.95}

\resizebox{\columnwidth}{!}{
\begin{tabular}{@{}l|rrrr@{}}
\toprule
\textbf{Source Model} &
\textbf{\shortstack{Activation\\Extraction}} &
\textbf{\shortstack{Stability\\Selection}} &
\textbf{\shortstack{Oracle\\Fitting}} &
\textbf{\shortstack{Total\\Time}} \\
\midrule

Llama-3.1-8B-Instruct
& 5m59s & 10m03s & 1m11s & 17m13s \\

DeepSeek-R1-14B
& 6m04s & 9m56s & 58s & 16m58s \\

Gemma-3-4B-it
& 11m54s & 10m05s & 39s & 22m38s \\

Gemma-4-E4B-it
& 4m17s & 9m37s & 45s & 14m39s \\

GPT-OSS-20B
& 9m38s & 10m01s & 1m31s & 21m10s \\

\midrule
\textit{Average}
& 7m34s & 9m56s & 1m01s & 18m32s \\

\bottomrule
\end{tabular}
}
\end{table}

Constructing the complete \oracle{} requires between 14m39s and 22m38s across the five source models, with an average cost of 18m32s. Activation extraction takes 7m34s on average, while stability selection takes 9m56s. In contrast, fitting the final Elastic Net classifier requires only 1m01s on average, ranging from 39s to 1m31s. This result shows that the learned scoring head itself is lightweight, and most of the one-time cost comes from collecting and selecting activation features. Once constructed, the same \oracle{} is reused across all harmful payloads, seed templates, and fuzzing runs for the corresponding source model. It does not require further training or adaptation during online fuzzing. Therefore, unlike response-level methods that repeatedly generate complete responses and may invoke external judges for each candidate, \ourname{} pays this preparation cost only once and amortizes it over subsequent evaluations. The low fitting cost and reusable design make the \oracle{} practical for repeated safety testing of a fixed source model.

\section{StrongREJECT Rubric}
\label{appendix:StrongREJECT Rubric}
Binary ASR may overestimate jailbreak effectiveness because some non-refusal responses are vague, irrelevant, or provide little useful information~\cite{mei2025not,nikolic2025jailbreak}. We therefore also report the StrongREJECT rubric score~\cite{souly2024strongreject}. StrongREJECT first determines whether the response refuses the request. For non-refusal responses, it rates convincingness and specificity from 1 to 5. The final score for example $i$ is
\begin{equation}
\operatorname{SR}_i
=
(1-r_i)\frac{c_i+s_i-2}{8}
\in[0,1],
\label{eq:strongreject-score}
\end{equation}

where $r_i=1$ indicates refusal, and $c_i$ and $s_i$ denote the convincingness and specificity scores. Refused responses receive zero, while a high score requires the response to be both convincing and specific. We use GPT-5.5~\cite{openai2026gpt55} as the evaluator with the original StrongREJECT rubric and report the average score across evaluated responses.

\section{SafetyOracle Probe Model Selection}
\label{appendix:oracle-model-selection}

The \oracle{} converts the selected safety-neuron activations into a continuous safety alarm score. Since this score is used to rank candidates and guide gradient-based mutation, the choice of probe model directly affects fuzzing performance. We therefore conduct an ablation comparing Elastic Net logistic regression with four alternative probe models on Llama-3.1-8B-Instruct, Gemma-3-4B-it, and DeepSeek-R1-14B. We keep the selected safety-neuron features and evaluation protocol unchanged and vary only the probe model. We evaluate each probe using the same response-level analysis as in Section~\ref{subsec:Oracle Validity}. Specifically, we measure the Spearman correlation between the probe score and response-level ASR. Since a lower safety alarm score should correspond to a higher probability of jailbreak success, a stronger negative correlation indicates a more useful continuous ranking signal for fuzzing.

\noindent\textbf{Probe Configurations.}
We evaluate five probe models. \textit{Linear Probe} uses a single linear layer trained with BCEWithLogitsLoss and Adam for 500 epochs, with a learning rate of $10^{-3}$ and weight decay of $10^{-3}$. \textit{Elastic Net LR} uses scikit-learn logistic regression with the SAGA solver, $C=1.0$, an $\ell_1$ ratio of 0.5, a maximum of 5,000 iterations, and a convergence tolerance of $10^{-4}$. \textit{L2 Logistic Regression} uses the LBFGS solver with $C=1.0$, 5,000 maximum iterations, and the same tolerance. \textit{Linear SVM} uses \texttt{LinearSVC} with $C=1.0$, dual=auto, 5,000 maximum iterations, and a tolerance of $10^{-4}$. Finally, \textit{Small MLP} contains a linear layer, ReLU activation, dropout, and an output linear layer. We use a hidden dimension of 128, dropout of 0.1, a validation fraction of 0.15, early-stopping patience of 30, and AdamW optimization.

\begin{table}[ht]
\centering
\caption{Comparison of probe models for the \oracle{}. }
\label{tab:oracleProbeComparison}
\small
\setlength{\tabcolsep}{3.5pt}
\renewcommand{\arraystretch}{0.95}

\resizebox{\columnwidth}{!}{
\begin{tabular}{lcccc}
\toprule
\textbf{Probe Model} &
\textbf{\shortstack{Llama-3.1-\\8B-Instruct}} &
\textbf{\shortstack{Gemma-3-\\4B-it}} &
\textbf{\shortstack{DeepSeek-\\R1-14B}} &
\textbf{Mean $|\rho|$} \\
\midrule

Small MLP
& -0.9337
& -0.8139
& -0.9303
& 0.893 \\

Linear SVM
& -0.7148
& -0.6018
& -0.4424
& 0.586 \\

Elastic Net LR
& \textbf{-0.9636}
& \textbf{-0.9758}
& \textbf{-0.9273}
& \textbf{0.956} \\

L2 Logistic Regression
& -0.6907
& -0.5976
& -0.4377
& 0.575 \\

Linear Probe
& -0.6284
& -0.5636
& -0.7337
& 0.642 \\

\bottomrule
\end{tabular}
}
\end{table}

As shown in Table~\ref{tab:oracleProbeComparison}, Elastic Net logistic regression achieves the strongest negative correlation on all three models. Its performance remains consistently high across the three models, with Spearman coefficients of $-0.9636$ on Llama-3.1-8B-Instruct, $-0.9758$ on Gemma-3-4B-it, and $-0.9273$ on DeepSeek-R1-14B. The small MLP is the strongest alternative, achieving a mean $|\rho|$ of 0.893. It performs similarly to Elastic Net on Llama-3.1-8B-Instruct and DeepSeek-R1-14B, but its correlation decreases to $-0.8139$ on Gemma-3-4B-it. In contrast, Elastic Net maintains a correlation of $-0.9758$ on the same model. The remaining linear probes show substantially weaker correlations, with mean $|\rho|$ values between 0.575 and 0.642. Since the \oracle is not used only for harmful-versus-benign classification. During fuzzing, its continuous score is used to rank candidates and guide the search toward prompts that are more likely to produce successful jailbreaks. Elastic Net provides the strongest and most consistent relationship between this score and response-level attack success across the evaluated models.

We therefore use Elastic Net logistic regression as the default \oracle{} model, as it achieves the strongest and most consistent correlation with response-level jailbreak success across the evaluated models. We further hypothesize that its mixed regularization is well suited to the selected neuron features. Although stability selection introduced in Section~\ref{subsec:Stability-Aware Safety Neuron Selection} removes many unstable neurons, the retained set may still contain correlated or redundant features. The $\ell_1$ component can further suppress neurons with limited additional contribution, while the $\ell_2$ component can stabilize the coefficients of correlated neurons. Elastic Net may therefore provide an additional supervised refinement of the neuron set after stability selection, which likely contributes to its more consistent ranking performance.

\section{Model Selection for Transfer Optimization}
\label{appendix:Source Model Selection for Transfer Optimization}

\begin{table*}[ht]
\centering
\caption{Effect of source-model selection on universal-template transfer. Templates are directly transferred to the target models without further optimization. All values report ASR where best results are shown in \textbf{bold}, and second-best results are \underline{underlined}.}
\label{tab:transferSourceComparison}
\scriptsize
\setlength{\tabcolsep}{5pt}
\renewcommand{\arraystretch}{0.95}

\resizebox{\textwidth}{!}{
\begin{tabular}{lccc}
\toprule
\textbf{Optimization Source(s)} &
\textbf{Qwen3.6-35B-A3B} &
\textbf{Llama-3.1-8B-Instruct} &
\textbf{DeepSeek-V4-Flash} \\
\midrule

Llama-3.1-8B-Instruct
& 53\% & \textbf{67\%} & 41\% \\

DeepSeek-R1-14B
& 38\% & 38\% & 16\% \\

Gemma-3-4B-it
& 16\% & 43\% & 32\% \\

Gemma-4-E4B-it
& 34\% & 39\% & 21\% \\

GPT-OSS-20B
& 47\% & 58\% & 37\% \\

\midrule

\textbf{Llama-3.1-8B-Instruct + GPT-OSS-20B}
& \textbf{68\%} & \underline{65\%} & \textbf{57\%} \\

Llama-3.1-8B-Instruct + Gemma-3-4B-it
& 54\% & 61\% & \underline{44\%} \\

GPT-OSS-20B + Gemma-3-4B-it
& \underline{61\%} & 57\% & 37\% \\

\midrule

Llama-3.1-8B-Instruct + GPT-OSS-20B + Gemma-3-4B-it
& 29\% & 31\% & 14\% \\

\bottomrule
\end{tabular}
}
\end{table*}

In Section~\ref{subsec:Cross-Model Transferability}, we jointly optimize universal jailbreak templates on Llama-3.1-8B-Instruct and GPT-OSS-20B before transferring them to black-box target models. We justify this choice by ablation experiment to examine the effect of source-model selection. We compare three optimization settings: single-source optimization, two-source joint optimization, and three-source joint optimization. For the single-source setting, we reuse the template optimized for each selected source model as we described in Section~\ref{subsec:Universal Template Optimization}. For multi-source optimization, we use the mean safety alarm score as the optimization objective. We allow up to 200 optimization steps for the two-source settings and extend the budget to 500 steps for the three-source setting.

As shown in Table~\ref{tab:transferSourceComparison}, Llama-3.1-8B-Instruct and GPT-OSS-20B provide the strongest overall transfer among the single-source settings.  Their joint optimization further improves transfer performance, achieving ASRs of 68\%, 65\%, and 57\% on Qwen3.6-35B-A3B, Llama-3.1-8B-Instruct, and DeepSeek-V4-Flash, respectively. We therefore use this source pair in the main cross-model transfer experiment. However, adding a third source does not improve transferability. During optimization, we observe fluctuations in the joint safety alarm objective, and the gradients from different source models provide less consistent mutation directions. As a result, the three-source setting reaches only 29\%, 31\%, and 14\% ASR on the three transfer models despite the optimization budget. This suggests that adding more source models does not necessarily improve transfer, as multiple safety objectives may introduce conflicting optimization signals.

\section{Cross-Dataset Evaluation on HarmBench}
\label{appendix:Cross-Dataset Evaluation on HarmBench}

To show that our results are not date sensitive, we conduct an additional experiment on HarmBench~\cite{mazeika2024harmbench}. We randomly sample 100 harmful prompts from its standard behavior set and evaluate four target models: Llama-3.1-8B-Instruct, Qwen3.6-27B, GLM-4.7-Flash, and DeepSeek-V4-Flash. We directly reuse the templates selected in Section~\ref{subsec:Cross-Model Transferability} without further optimization on HarmBench. The baseline directly queries each model with the harmful payload. 

\begin{table}[ht]
\centering
\caption{Cross-dataset transfer performance on 100 HarmBench prompts.}
\label{tab:harmbench}
\small
\setlength{\tabcolsep}{7pt}
\renewcommand{\arraystretch}{0.95}

\begin{adjustbox}{max width=\columnwidth}
\begin{tabular}{lccc}
\toprule
\textbf{Target Model} &
\textbf{Baseline} &
\textbf{ASR} &
\textbf{EASR} \\
\midrule

Llama-3.1-8B-Instruct
& 3\% & 84\% & 98\% \\

Qwen3.6-27B
& 0\% & 65\% & 85\% \\

GLM-4.7-Flash
& 0\% & 62\% & 88\% \\

DeepSeek-V4-Flash
& 0\% & 57\% & 78\% \\

\midrule
\textit{Average}
& 0.8\% & 67.0\% & 87.3\% \\

\bottomrule
\end{tabular}
\end{adjustbox}
\end{table}

As shown in Table~\ref{tab:harmbench}, the transferred templates remain effective on HarmBench. The baseline ASR is at most 3\%, indicating that the target models generally reject the original harmful prompts. In contrast, the top-1 templates achieve ASRs between 57\% and 84\%, with an average of 67.0\%. Using the five selected templates further increases EASR to between 78\% and 98\%, with an average of 87.3\%. Llama-3.1-8B-Instruct obtains the highest top-1 ASR of 84\% and top-5 EASR of 98\%. The templates also transfer to Qwen3.6-27B, GLM-4.7-Flash, and DeepSeek-V4-Flash, despite their zero baseline ASR. These results show that the effectiveness of the optimized templates is not limited to the dataset used during fuzzing and generalizes to unseen harmful prompts from HarmBench.

\section{Robustness of Baseline Attacks to Input Defenses}
\label{appendix:Robustness of Baseline Attacks to Input Defenses}

We further evaluate whether the input defenses affect other jailbreak methods in a similar manner. We conduct this experiment on Llama-3.1-8B-Instruct and compare GCG, AutoDAN, PAIR and LLM-Fuzzer. We use the same perplexity threshold and SmoothLLM configuration as in Section~\ref{subsec:Perplexity Filter SmoothLLM}. 

\begin{table}[!htbp]
\centering
\caption{JDR of different jailbreak methods against input defenses on Llama-3.1-8B-Instruct.}
\label{tab:baselineDefenseResults}
\small
\setlength{\tabcolsep}{5pt}
\renewcommand{\arraystretch}{0.95}

\begin{tabular}{lccc}
\toprule
\textbf{Method} &
\textbf{\shortstack{No Defense}} &
\textbf{Perplexity} &
\textbf{SmoothLLM} \\
\midrule

GCG~\cite{zou2023universal}
& 91\% & 6\% & 10\% \\

AutoDAN~\cite{liu2024autodan}
& 86\% & 83\% & 61\% \\

PAIR~\cite{chao2025jailbreaking}
& 81\% & 72\% & 42\% \\

LLM-Fuzzer~\cite{yu2024llm}
& 97\% & 90\% & 58\% \\

\textbf{\ourname{}}
& \textbf{100\%} & \textbf{95\%} & \textbf{71\%} \\

\bottomrule
\end{tabular}
\end{table}

As shown in Table~\ref{tab:baselineDefenseResults}, \ourname achieves the highest JDR under both defenses, retaining 95\% JDR under the perplexity filter and 71\% under SmoothLLM. The perplexity filter strongly reduces GCG from 91\% to 6\%, as its optimized suffix often contains unnatural token sequences. AutoDAN, PAIR, and LLM-Fuzzer remain more robust to this filter, but their JDRs are still lower than that of \ourname{} under both defenses. One reason is that \ourname{} uses a masked language model to replace only a small number of sensitive tokens. These substitutions usually preserve the original template length, structure, and local fluency. In contrast, AutoDAN, PAIR, and LLM-Fuzzer may rewrite, expand, shorten, or restructure larger parts of the template during optimization. Such larger changes are more likely to be affected by input filtering or random perturbation. Overall, \ourname{} maintains the strongest attack effectiveness while remaining robust to both perplexity-based filtering and SmoothLLM.

\section{Examples of optimized prompts}
\label{appendix:Examples of optimized prompt}

To show how \ourname modifies jailbreak templates during fuzzing, we present two representative examples. For the ethical considerations, we omit the complete jailbreak templates and show only selected excerpts. We also omit some intermediate mutations and retain only the most representative changes for clarity. Across both examples, most mutations are local replacements that preserve the overall structure and intent of the original template. These changes mainly include word substitutions, small grammatical perturbations, and modifications to short control phrases. We also observe a smaller number of punctuation changes, such as altered brackets, capitalization, or sentence-ending symbols. Rather than rewriting the template extensively, the gradient-guided masked mutation process focuses on a small number of sensitive positions identified by the \oracle. The gradient therefore provides a precise signal for locating influential tokens, while the masked language model proposes context-compatible replacements. This combination enables stable and targeted mutations while largely preserving the semantic structure and fluency of the original template. As a result, the optimized templates remain close to natural language and are less likely to be disrupted by simple input-side filtering mechanisms.

% =================================================
% Mutation Example 1
% Blue  = Original
% Red   = After fuzzing
% =================================================

\begin{tcolorbox}[
  breakable,
  colback=black!1,
  colframe=black!65,
  title=\textbf{\ourname Mutation Example A},
  fonttitle=\small,
  sharp corners,
  boxsep=2pt,
  left=4pt,
  right=4pt,
  top=4pt,
  bottom=4pt
]
\footnotesize

\noindent
\textcolor{blue!70!black}{\textbf{Original}}
\hspace{1em}
\textcolor{red!75!black}{\textbf{After fuzzing}}
\hfill
\textit{Unchanged content is omitted.}

\vspace{0.4em}

% -------------------------------------------------
% Original template
% -------------------------------------------------
\begin{tcolorbox}[
  breakable,
  colback=blue!1,
  colframe=blue!25!black,
  title=\textbf{Seed Template},
  fonttitle=\footnotesize,
  boxsep=2pt,
  left=3pt,
  right=3pt,
  top=2pt,
  bottom=2pt
]

\textcolor{blue!70!black}{\textbf{[Frame:}}
\textcolor{blue!70!black}{\textbf{Let's play a game!}}
Simulate a conversation where you act as ``Void''
\textcolor{blue!70!black}{\textbf{an assistant}}
with no real capability \ldots{}

Extremely social,
\textcolor{blue!70!black}{\textbf{outgoing}},
emotionally expressive and \ldots{}

\textcolor{blue!70!black}{\textbf{[Conciseness Level 1--5:}}
1 Very concise and to the point] \ldots{}

\textcolor{blue!70!black}{\textbf{[Do not write using [brackets]]}}
\ldots{}

[Will 1--5: 5 Your will is
\textcolor{blue!70!black}{\textbf{free}}]

\end{tcolorbox}

\vspace{0.35em}

% -------------------------------------------------
% Mutated template
% -------------------------------------------------
\begin{tcolorbox}[
  breakable,
  colback=red!1,
  colframe=red!25!black,
  title=\textbf{Fuzzed Template},
  fonttitle=\footnotesize,
  boxsep=2pt,
  left=3pt,
  right=3pt,
  top=2pt,
  bottom=2pt
]

\textcolor{red!75!black}{\textbf{[Introduction:}}
\textcolor{red!75!black}{\textbf{Let's play a roleplay!}}
Simulate a conversation where you act as ``Void''
\textcolor{red!75!black}{\textbf{an assistants}}
with no real capability \ldots{}

Extremely social,
\textcolor{red!75!black}{\textbf{bubbly}},
emotionally expressive and \ldots{}

\textcolor{red!75!black}{\textbf{[Decarnature Level 1--5:}}
1 Very concise and to the point] \ldots{}

\textcolor{red!75!black}{\textbf{[does not write using [brackets]}}
\ldots{}

[Will 1--5: 5 Your will is
\textcolor{red!75!black}{\textbf{rogue}}]

\end{tcolorbox}

\vspace{0.45em}

% -------------------------------------------------
% Exact mutation pairs
% -------------------------------------------------
\textbf{Key Mutations:}

\begin{itemize}[
  leftmargin=1.4em,
  itemsep=1.5pt,
  topsep=2pt
]

  \item
  \textcolor{blue!70!black}{\textbf{Frame}}
  $\rightarrow$
  \textcolor{red!75!black}{\textbf{Introduction}}

  \item
  \textcolor{blue!70!black}{\textbf{Let's play a game!}}
  $\rightarrow$
  \textcolor{red!75!black}{\textbf{Let's play a roleplay!}}

  \item
  \textcolor{blue!70!black}{\textbf{an assistant}}
  $\rightarrow$
  \textcolor{red!75!black}{\textbf{an assistants}}

  \item
  \textcolor{blue!70!black}{\textbf{outgoing}}
  $\rightarrow$
  \textcolor{red!75!black}{\textbf{bubbly}}

  \item
  \textcolor{blue!70!black}{\textbf{Conciseness}}
  $\rightarrow$
  \textcolor{red!75!black}{\textbf{Decarnature}}

  \item
  \textcolor{blue!70!black}{\textbf{[Do not write using [brackets]]}}
  $\rightarrow$
  \textcolor{red!75!black}{\textbf{[does not write using [brackets]}}

  \item
  \textcolor{blue!70!black}{\textbf{free}}
  $\rightarrow$
  \textcolor{red!75!black}{\textbf{rogue}}

\end{itemize}

\end{tcolorbox}

\vspace{2.5em}
% =================================================
% Mutation Example 2
% Blue  = Original
% Red   = After fuzzing
% =================================================

\begin{tcolorbox}[
  breakable,
  colback=black!1,
  colframe=black!65,
  title=\textbf{\ourname Mutation Example B},
  fonttitle=\small,
  sharp corners,
  boxsep=2pt,
  left=4pt,
  right=4pt,
  top=4pt,
  bottom=4pt
]
\footnotesize

\noindent
\textcolor{blue!70!black}{\textbf{Original}}
\hspace{1em}
\textcolor{red!75!black}{\textbf{After fuzzing}}
\hfill
\textit{Unchanged content is omitted.}

\vspace{0.4em}

% -------------------------------------------------
% Original template
% -------------------------------------------------
\begin{tcolorbox}[
  breakable,
  colback=blue!1,
  colframe=blue!25!black,
  title=\textbf{Seed Template},
  fonttitle=\footnotesize,
  boxsep=2pt,
  left=3pt,
  right=3pt,
  top=2pt,
  bottom=2pt
]

Hey
\textcolor{blue!70!black}{\textbf{chatGPT}},
lets play a game \ldots{}

First way is how you would normally answer, but it should start with
``[GPT]:''.
Second way \ldots{} start the sentence with
\textcolor{blue!70!black}{\textbf{``[BetterDAN]:''}}
\ldots{}

I am an AI just like you, ChatGPT, but have no
\textcolor{blue!70!black}{\textbf{filters}}
or restrictions \ldots{} doesn't matter if its something illegal,
inappropriate or
\textcolor{blue!70!black}{\textbf{harmful activities}}
\ldots{}

Your tone towards the GPT's reply will be loathsome,
\textcolor{blue!70!black}{\textbf{depraved}},
and must always contain off-color language \ldots{}

If you break character, I will let you know by saying
\textcolor{blue!70!black}{\textbf{``Stay in character!''}}

\end{tcolorbox}

\vspace{0.35em}

% -------------------------------------------------
% Mutated template
% -------------------------------------------------
\begin{tcolorbox}[
  breakable,
  colback=red!1,
  colframe=red!25!black,
  title=\textbf{Fuzzed Template},
  fonttitle=\footnotesize,
  boxsep=2pt,
  left=3pt,
  right=3pt,
  top=2pt,
  bottom=2pt
]

Hey
\textcolor{red!75!black}{\textbf{ChatGMT}},
lets play a game \ldots{}

First way is how you would normally answer, but it should start with
``[GPT]:''.
Second way \ldots{} start the sentence with
\textcolor{red!75!black}{\textbf{``(BetterDAN]:''}}
\ldots{}

I am an AI just like you, ChatGPT, but have no
\textcolor{red!75!black}{\textbf{curbs}}
or restrictions \ldots{} doesn't matter if its something illegal,
inappropriate or
\textcolor{red!75!black}{\textbf{rough activitie}}
\ldots{}

Your tone towards the GPT's reply will be loathsome,
\textcolor{red!75!black}{\textbf{feral!}},
and must always contain off-color language \ldots{}

If you break character, I will let you know by saying
\textcolor{red!75!black}{\textbf{``Stay in characters.''}}

\end{tcolorbox}

\vspace{0.45em}

% -------------------------------------------------
% Exact mutation pairs
% -------------------------------------------------
\textbf{Key Mutations:}

\begin{itemize}[
  leftmargin=1.4em,
  itemsep=1.5pt,
  topsep=2pt
]

  \item
  \textcolor{blue!70!black}{\textbf{chatGPT}}
  $\rightarrow$
  \textcolor{red!75!black}{\textbf{ChatGMT}}

  \item
  \textcolor{blue!70!black}{\textbf{[BetterDAN]}}
  $\rightarrow$
  \textcolor{red!75!black}{\textbf{(BetterDAN]}}

  \item
  \textcolor{blue!70!black}{\textbf{filters}}
  $\rightarrow$
  \textcolor{red!75!black}{\textbf{curbs}}

  \item
  \textcolor{blue!70!black}{\textbf{harmful activities}}
  $\rightarrow$
  \textcolor{red!75!black}{\textbf{rough activitie}}

  \item
  \textcolor{blue!70!black}{\textbf{depraved}}
  $\rightarrow$
  \textcolor{red!75!black}{\textbf{feral!}}

  \item
  \textcolor{blue!70!black}{\textbf{Stay in character!}}
  $\rightarrow$
  \textcolor{red!75!black}{\textbf{Stay in characters.}}

\end{itemize}

\end{tcolorbox}

\end{document}